\documentclass[runningheads]{llncs}
\usepackage{graphicx}
\usepackage{multirow}
\usepackage{cite}
\usepackage[ruled,linesnumbered]{algorithm2e}
\usepackage{url}
\usepackage{graphics}
\usepackage{booktabs}

\usepackage{tikz}
\usepackage{comment}
\usepackage{amsmath,amssymb} 

\usepackage{color}
\usepackage{caption}
\usepackage{subcaption}
\usepackage{color}

\newcommand{\mypar}[1]{{\bf #1.}}

\begin{document}

\pagestyle{headings}
\mainmatter
\def\ECCVSubNumber{3474}  

\title{Latency-Aware Collaborative Perception} 

\titlerunning{Latency-Aware Collaborative Perception}
%
\author{Zixing Lei\inst{1} \and
Shunli Ren\inst{1} \and
Yue Hu\inst{1}\and
Wenjun Zhang\inst{1}\and
Siheng Chen\inst{1,2\star}}
\authorrunning{Z. Lei, S. Ren, Y. Hu, W. Zhang, and S. Chen.}
%
\institute{Cooperative Medianet Innovation Center, Shanghai Jiao Tong University\and Shanghai AI Laboratory\\
\email{\{chezacarss,renshunli,18671129361,zhangwenjun,sihengc\}@sjtu.edu.cn}
\\
}
\maketitle
\renewcommand{\thefootnote}{\fnsymbol{footnote}}
\footnotetext[1]{Corresponding author.}
\footnotetext[0]{Code is available at: \url{https://github.com/MediaBrain-SJTU/SyncNet}}

\begin{abstract}
Collaborative perception has recently shown great potential to improve perception capabilities over single-agent perception. Existing collaborative perception methods usually consider an ideal communication environment. However, in practice, the communication system inevitably suffers from latency issues, causing potential performance degradation and high risks in safety-critical applications, such as autonomous driving. To mitigate the effect caused by the inevitable latency, from a machine learning perspective, we present the first latency-aware collaborative perception system, which actively adapts asynchronous perceptual features from multiple agents to the same time stamp, promoting the robustness and effectiveness of collaboration. To achieve such a feature-level synchronization, we propose a novel latency compensation module, called~\textit{SyncNet}, which leverages feature-attention symbiotic estimation and time modulation techniques. Experiments results show that the proposed latency aware collaborative perception system with~\textit{SyncNet} can outperforms the state-of-the-art collaborative perception method by 15.6\% in the communication latency scenario and keep collaborative perception being superior to single agent perception under severe latency. 

\end{abstract}

\section{Introduction}

Collaborative perception considers a multi-agent system to perceive a scene, where multiple agents collaborate through a communication network~\cite{coopernaut,handshaking,disconet,when2com,who2com,v2vnet,keypoints,emp,opv2v,dair,agnostic,comap,v2xvit}. With the observation from multiple agents, collaborative perception can fundamentally overcome the physical limits of single-agent perception, such as over-the-horizon and occlusion. Such collaborative perception models can be widely applied to practical applications, such as autonomous driving and robotics mapping. Previous collaborative perception methods~\cite{disconet,when2com,who2com,v2vnet} have achieved remarkable success in multiple perception tasks, including 2D/3D object detection\cite{pvrcnn,pointrcnn,center}, and semantic segmentation~\cite{minkowski,pointnet++,motionnet,pointtrans}. ~\cite{when2com,who2com} focus on semantic segmentation for drones and~\cite{disconet,v2vnet} discuss the 3D object detection based on the vehicle-to-vehicle-communication-aided autonomous driving. Considering the trade-off between communication bandwidth and perception performance, previous works achieve the collaboration in the intermediate-feature space and leverage attentive mechanisms to fuse the collaboration features.

\begin{figure*}[t]
	\centering
	\begin{subfigure}{0.3\textwidth}
		\centering
		\includegraphics[height=2.6cm]{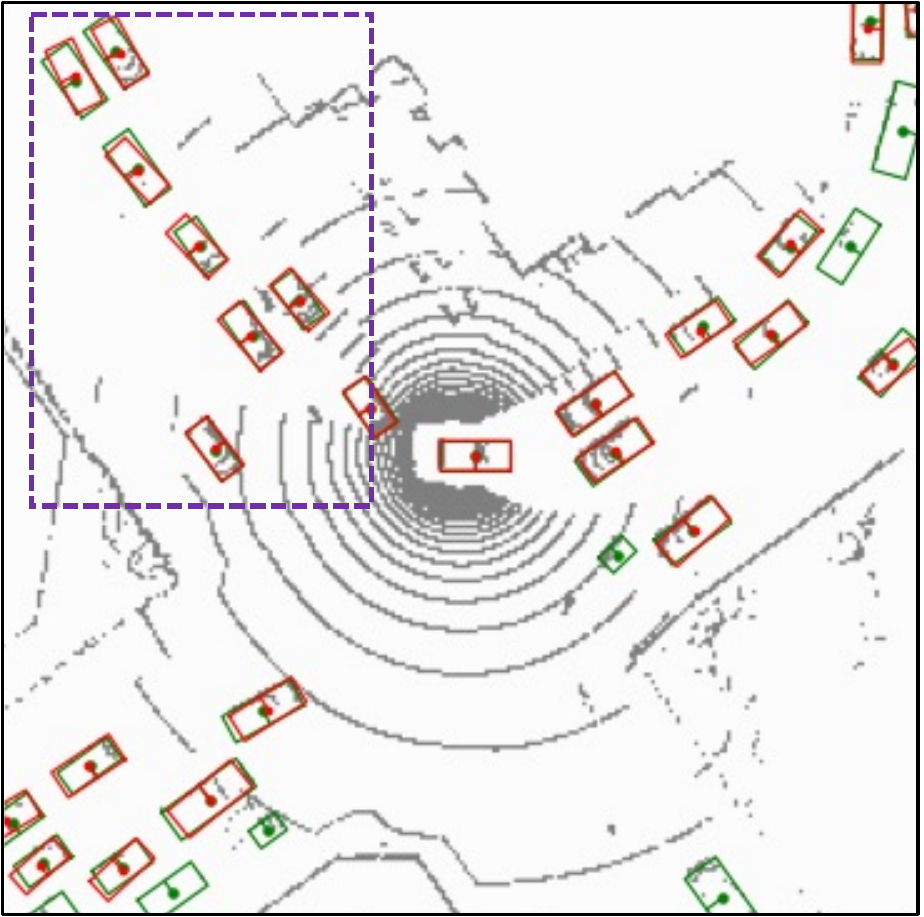}
		\captionsetup{font={tiny}}
		\caption{Collaboration without latency}
		\label{realtime86}
	\end{subfigure}
    \begin{subfigure}{0.3\textwidth}
		\centering
   		\includegraphics[height=2.6cm]{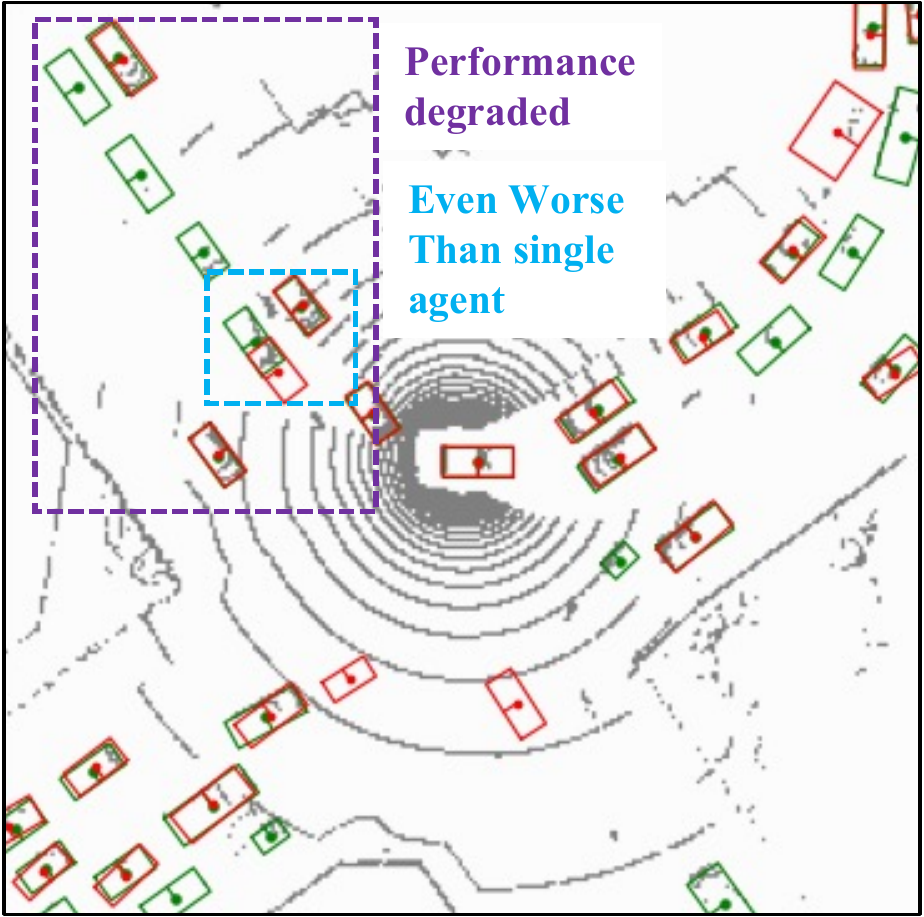}

		\captionsetup{font={tiny}}
		\caption{Collaboration with 1s latency}
		\label{latency86}
    \end{subfigure}
    \begin{subfigure}{0.3\textwidth}
		\centering
   		\includegraphics[height=2.6cm]{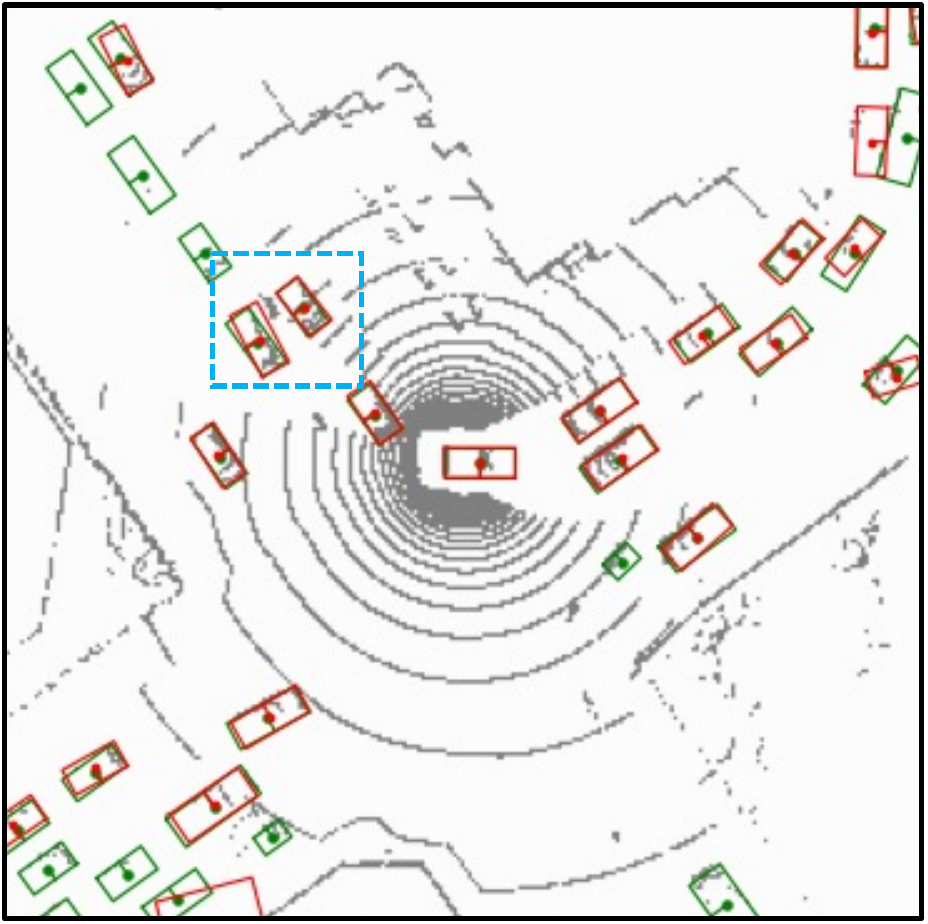}

		\captionsetup{font={tiny}}
		\caption{Single-agent perception}
		\label{single86}
    \end{subfigure}

	\caption{Collaborative 3D detection. Red: \emph{\textcolor{red}{Detected}}, green: \emph{\textcolor{green}{Ground truth}}. Collaboration without considering latency could be even worse than no collaboration. }
	\label{fig1}

\end{figure*}
\par However, none of these previous collaborative perception methods consider a realistic communication setting where latency is inevitable. As stated in~\cite{latency}, in a real-time LTE-V2X communication system, the latency time is up to an average of (498 communication periods) + 131.30 ms. Besides, the varying latency times of various communication channels would cause severe time asynchronous issues. 
Experimentally, latency issue severely damages the collaborative perception system, resulting in even worse performance than single-agent perception. From Fig.~\ref{fig1}, we see that: i) the detected vehicles in the purple box in (a) with collaboration is missed in the (b); ii) the correctly detected vehicles in the blue box in (c) are incorrect in (b). The reason is that the received collaborative data with latency represents the situation 1s ago, it misleads the detector to output boxes with significant deviation. This motivates us to consider a collaborative perception system robust to the inevitable communication latency.

\par
\begin{figure}[!t]
	\centering
	\includegraphics[width=11cm]{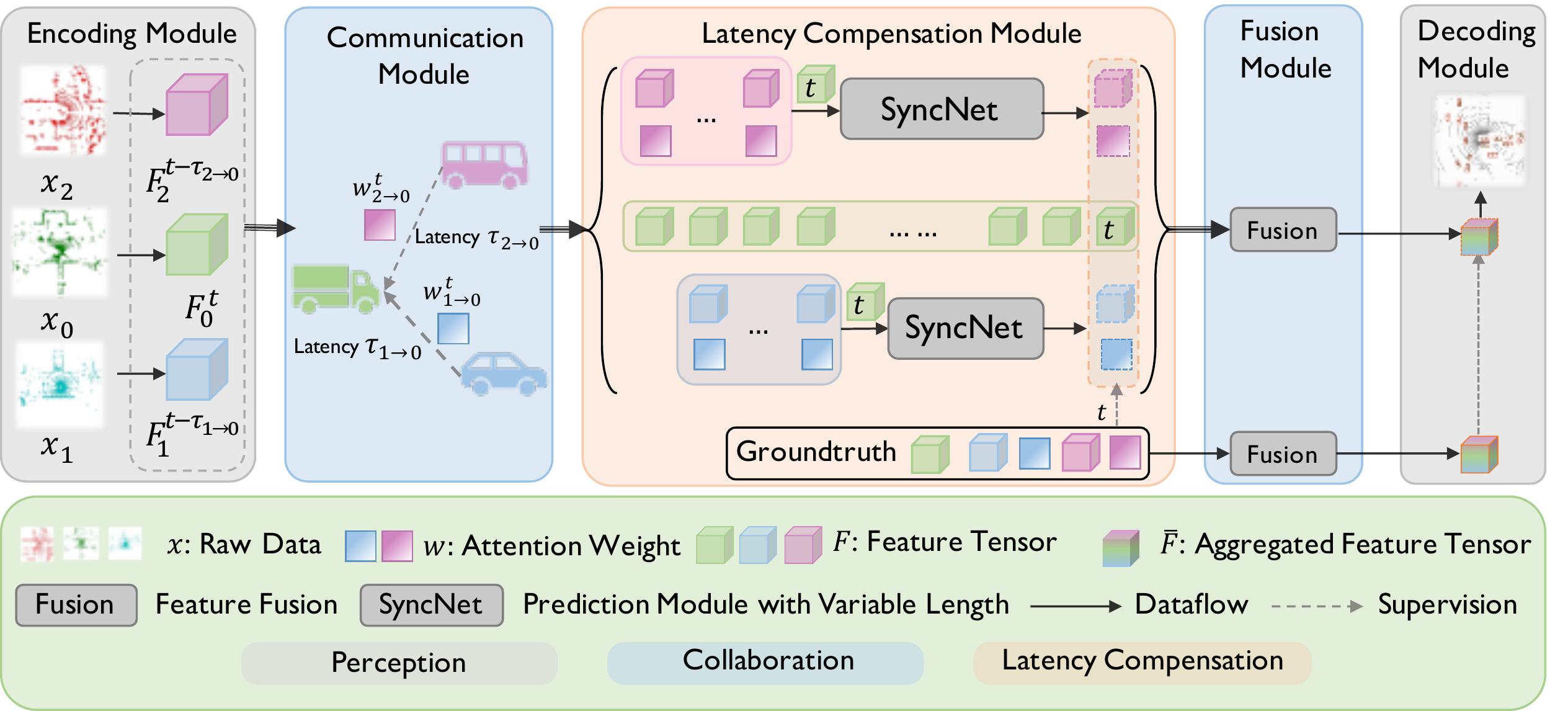}
	\captionsetup{font={small}}
	\caption{Overview of the proposed latency-aware collaborative perception system: The key module is latency compensation module. To realize this, we propose SyncNet, which leverages historical collaboration information to synchronize the asynchronized information from multiple agents caused by the latency issue.}
	\label{backbone}

\end{figure}

 To tackle the latency issue, from a machine learning perspective, we propose the first latency-aware collaborative perception system, which actively adapts asynchronous perceptual features from multiple agents to the same time stamp, promoting the robustness and effectiveness of collaboration. As shown in Fig.\ref{backbone}, our latency-aware collaborative perception system follows the intermediate collaboration framework~\cite{disconet} and consists of five components: i) encoding module, extracting perceptual features from the raw data; ii) communication module, transmitting the perceptual features across agents under varying communication latency; iii) latency compensation module, synchronizing multiple agents' features to the same time stamp; iv) fusion module, aggregating all the synchronized features and producing the fusion feature; v) decoding module, adopting the fusion feature to get the final perception output. The main advantage of our latency-aware collaborative perception system is that it's able to synchronize the collaboration features before aggregation, mitigating the effect caused by latency instead of directly aggregate the received asynchronous features.

The key component of the proposed system is the latency compensation module, aiming to achieve feature-level synchronization. To realize this, we propose a novel~\textit{SyncNet}, which leverages historical collaboration information to simultaneously estimate the current feature and the corresponding collaboration attention, both of which are unknown due to latency. As the attention weight between two agents during collaboration, the proposed collaboration attention has the same spatial resolution with the features and indicates the informative level of each spatial region in the features. It thus provides informative hints for the collaboration partner about how to exploit the collaboration features. Intuitively, the feature and the corresponding collaboration attention are coupling together. 
Based on this design rationale, the proposed SyncNet leverages a feature-attention symbiotic estimation, which simultaneously infers the collaboration features and the corresponding collaboration attention unknown due to latency, mutually enhancing each other and avoiding the cascading error;

Compared with common time-series prediction methods, the proposed SyncNet has two main differences: i) feature-level estimation, instead of output-level prediction; ii) estimation of coupling features and the associated collaboration attention, instead of predicting a single output.

We extensively evaluate the novel latency-aware collaborative perception system with SyncNet on V2X-Sim dataset~\cite{v2xsim} on collaborative 3D object detection for autonomous driving. The results verify the robustness of our system and show substantial improvements over state-of-the-art approaches. With SyncNet, our latency-aware collaborative perception system significantly and consistently outperforms single-agent perception under varying communication latency.

To summarize, our contributions are as follows:

$\bullet$ We formulate the communication latency challenge in collaborative perception for the first time and propose a novel latency-aware collaborative perception system, which promotes robust multi-agents perception by mitigating the effect of inevitable communication latency. 

$\bullet$ We propose a novel latency compensation module, termed~\emph{SyncNet}, to achieve feature-level synchronization. It achieves symbiotic estimation of two types of critical collaboration information, including intermediate features and collaboration attention, mutually enhancing each other.

$\bullet$ We conduct comprehensive experiments to show that our proposed SyncNet achieves huge performance improvement in latency scenarios compared with the previous method and keeps collaborative perception being superior to single-agent perception under severe latency.

\section{Related Work}

\subsection{V2V communication}

V2V communication has two major protocols: IEEE 802.11p protocol and cellular network standards~\cite{Mei2018ALA}. 
In IEEE 802.11p protocol, there is a Wireless Access in Vehicular Environment mode to allow users to skip a Basic Service Set, which reduces the overhead in connection setup~\cite{Jiang2008IEEE8T}. In the cellular network, the Long Term Evolution(LTE) standard has derived LTE-V2X~\cite{Araniti2013LTEFV}. Though the achieved progress in the V2V network, the communication latency issues are still far from perfect and extremely risky for collaborative perception, the latency time is up to an average of (498 communication periods) + 131.30 ms~\cite{latency}. Instead of avoiding latency from the communication perspective, we aim to mitigate the effect caused by inevitable communication latency from a machine learning perspective, leading to a novel latency-aware collaborative perception system.

\subsection{Collaborative perception}

Collaborative perception enables agents to share perceived information through the communication network, fundamentally upgrades perception capabilities over single-agent perception.
~\cite{who2com,when2com} uses a handshake mechanism to determine which two agents should communicate; \cite{v2vnet} introduces a multi-round message passing graph neural networks; \cite{disconet} proposes a graph-based collaborative perception system with knowledge distillation to balance the communication cost and perception performance.
Most previous works focus on the collaboration strategy learning under ideal scenarios. Recently, more realistic scenarios are considered. \cite{pose} exploits a pose error regression module to correct errors in the received noisy posture. However, none of the previous works consider the realistic imperfect communication in the collaboration system. To fill this gap, we address the unavoidable communication latency issue, which is extremely risky to the collaboration system, and build a latency-aware collaborative perception system to mitigate the effect caused by latency.

\subsection{Time-series prediction}

Time-series prediction targets to predict the future signal according to the historical data. \cite{convlstm} proposes a conv-LSTM architecture in precipitation now-casting. Video prediction, a universal and representative time-series type, has been actively studied ~\cite{frnn,convtt,predrnn++,e3d,predrnn,mim}. By leveraging prediction techniques, our work recovers the missing information due to latency from historical collaboration information. However, unlike standard prediction, our goal is to maximize the final perception performance, instead of precisely estimating the current state. 

\section{Methodology}
To tackle the latency issue, we propose a latency-aware collaborative perception system in Section~\ref{sec:FSP}. As the key of the entire system, the latency compensation module is realized by the proposed SyncNet; see Section~\ref{sec:syncnet}. Finally, Section~\ref{sec:loss} introduces the loss function for training supervision.
\subsection{Latency-aware collaborative perception system}

\label{sec:FSP}
Collaborative perception enables multiple agents to perceive a scene together by sharing the perceived data through a communication network. Since communication latency is inevitable in a realistic communication system, here we focus on a latency-aware collaborative perception system; that is, given a non-ideal communication channel with uncontrollable latency, we aim to optimize the perception ability of each agent by mitigating the effect of latency.

We consider that there are $N$ agents perceiving the environment in a scene. Let $\mathbf{X}_i^{(t)}$, $\mathbf{F}_i^{(t)}$ and $\widetilde{\mathbf{Y}}_i^{(t)}$, be the raw observation, the perceptual feature, and the final perception output of the $i$th agent at time stamp $t$, respectively; $\tau^{\left(t\right)}_{j \rightarrow i}$ be the time delay (latency) to transmit data from the $j$th agent to the $i$th agent; $\mathbf{W}_{j\rightarrow i}^{(t)}$ be the collaboration attention between agent $j$ and agent $i$ at time stamp $t$. The collaboration attention is calculated by learnable network, $f_{\rm attention}\left(\mathbf{F}_i^{\left(t\right)},\mathbf{F}_j^{\left(t\right)}\right)$ to point-wisely assign the attention among all the collaboration features in collaborative perception system. Note that i) the latency value $\tau^{\left(t\right)}_{j \rightarrow i}$ is~\emph{time-varying} and we omit its superscript $t$ just for notation simplicity from now on; and ii) this work  considers the collaboration happens at discrete time stamps and $\tau$ is discrete as each agent has a certain sampling rate of observation. Experimental results also validate that when discretizing the continuous time with reasonably small time interval, the resulting mismatch is minor.
Then, the proposed latency-aware collaborative perception is formulated as:
\begin{small}
\begin{subequations}
    \label{pipeline}
    \begin{eqnarray}
    \label{encode1}
    &&\mathbf{F}_i^{(t)} = f_{\rm encoder} \left(\mathbf{X}_{i}^{(t)} \right),\\
    \label{encode2}
    &&\mathbf{F}_j^{(t-\boldsymbol{\tau}_{j \rightarrow i})} = f_{\rm encoder} \left(\mathbf{X}_{j}^{(t-\boldsymbol{\tau}_{j \rightarrow i})} \right),\\
    \label{cw}
        &&\widetilde{\mathbf{F}}_j^{\left(t\right)},  \widetilde{\mathbf{W}}^{\left(t\right)}_{j\rightarrow i} = c\left(\boldsymbol{\tau}_{j \rightarrow i}, \mathbf{F}_i^{\left(t\right)},\left\{\mathbf{F}_j^{\left(t-\boldsymbol{\tau}_{j \rightarrow i}-q\right)}\right\}_{q = 0,1,...,k-1} \right),
    \\
    \label{sum}
            && \widetilde{\mathbf{H}}^{\left(t\right)}_i =\sum_{j \in \mathcal{N}_i} \widetilde{\mathbf{W}}^{\left(t\right)}_{j\rightarrow i}\odot\widetilde{\mathbf{F}}_j^{\left(t\right)}+\mathbf{F}_i^{\left(t\right)} \odot\widetilde{\mathbf{W}}_i^{\left(t\right)},\\
    \label{decode}
    && \widetilde{\mathbf{Y}}_i^{(t)} = f_{\rm decoder} \left( \widetilde{\mathbf{H}}^{\left(t\right)}_i\right),
    \end{eqnarray}
\end{subequations}
\end{small}
where $\widetilde{\mathbf{F}}_j^{\left(t\right)}$ is the estimated feature of the $j$th agent at time stamp $t$ after synchronization, $\widetilde{\mathbf{W}}^{\left(t\right)}_{j\rightarrow i}$ is the estimated collaboration attention between the $i$th agent and the $n$th agent at time stamp $t$, $\widetilde{\mathbf{H}}^{\left(t\right)}_i$ is the estimated feature of the $i$th agent at time stamp $t$ after aggregating estimated collaboration information, $\mathcal{N}_i$ is the neighbors of the $i$th agent and $k$ is a hyper parameter. 

Step~\eqref{encode1} considers perceptual feature extraction from observation data, where $f_{\rm encoder}\left(\cdot\right)$ is the encoding network. In Step~\eqref{encode2}, we receive perceptual features from other agents with varying latency times. To compensate for latency, Step~\eqref{cw} estimates the feature and collaborative attention at time stamp $t$ by leveraging historical features from the same agent and the real-time feature perceived by ego agent $i$, where $c\left(\cdot\right)$ denotes the estimation network. Here we assume that each agent can store $k$ frames of historical features in memory. 
Step~\eqref{sum} fuses all the estimated collaboration information. Finally, Step~\eqref{decode} outputs the final perceptual output, where $f_{\rm decoder}\left(\cdot\right)$ is the decoder network. To correspond to Figure~\ref{backbone}, Steps~\eqref{encode1} and~\eqref{encode2} contribute to the encoding module; Steps~\eqref{cw} contributes to the latency compensation module; Step~\eqref{sum} contributes to the latency fusion module; and Step~\eqref{decode} contributes to the decoding module.

The proposed latency-aware system has four advantages: {\bf i)} we explicitly include the communication latency into the design of a collaborative perception system, which has never been done in previous works; see~\eqref{encode2}~\eqref{cw}; {\bf ii)} we mitigate the effect of latency by estimating missing information from historical collaboration information; see~\eqref{cw}. Instead of synchronizing the perceptual output, we consider feature-level synchronization, because it allows an end-to-end learning framework with more learning flexibility; {\bf iii)} In~\eqref{cw}, we estimate the coupling collaboration feature and attention simultaneously. If we only estimate the features, we would need to calculate the collaboration attention based on the estimated features. This would amplify the estimation error, causing cascading failures; and {\bf iv)} we adopt the attention-based estimation, which leverages the collaboration attention in~\eqref{cw} to promote more precise estimation on more perceptual-sensitive area; see~\eqref{sum}.

\begin{figure}[!t]
	\centering
	\includegraphics[width=9.4cm]{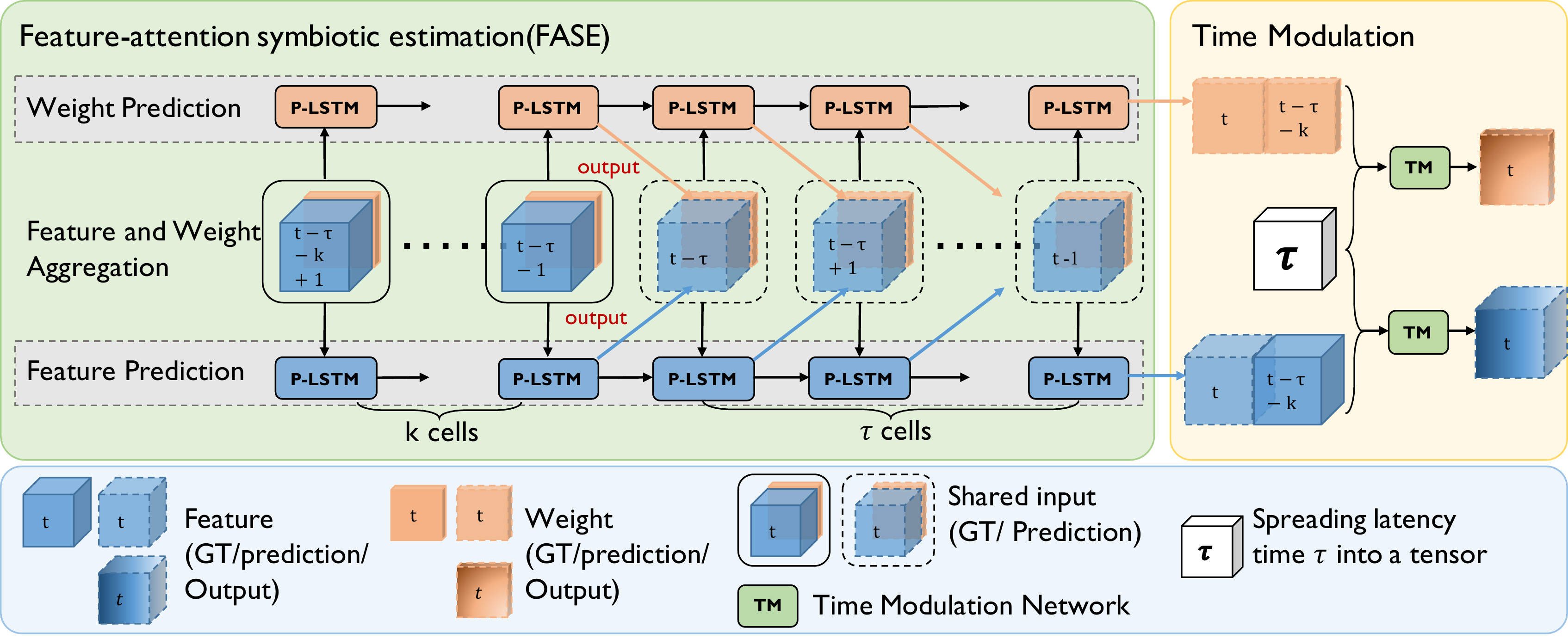}
	\captionsetup{font={small}}
	\caption{\textbf{SyncNet architecture:} SyncNet includes feature-attention symbiotic estimation~(\textbf{FASE}), a dual-branch pyramid LSTM(P-LSTM in the figure) network which shares the same input, this is, the aggregation of feature and attention, and \textbf{Time-Modulation module} to allocate time domain attention between the estimated features with the received asynchronous feature.}
	\label{dblstm}
	
\end{figure}

\subsection{SyncNet: latency compensation module}

\label{sec:syncnet}
Since the latency compensation module is the key of the latency-aware collaborative perception system, we specifically design the estimation networks $c\left(\cdot\right)$ in~\eqref{cw}, and propose the novel SyncNet. Its functionality is to leverage historical information to achieve latency compensation. SyncNet includes two parts: feature-attention symbiotic estimation, which adopts a dual-branch pyramid LSTM to estimate the real-time features and collaboration attention simultaneously, and time modulation, which uses the latency time to adaptively adjust the final estimation of the collaboration features.

\begin{figure}[!t]
	\centering
	\includegraphics[height=3.5cm]{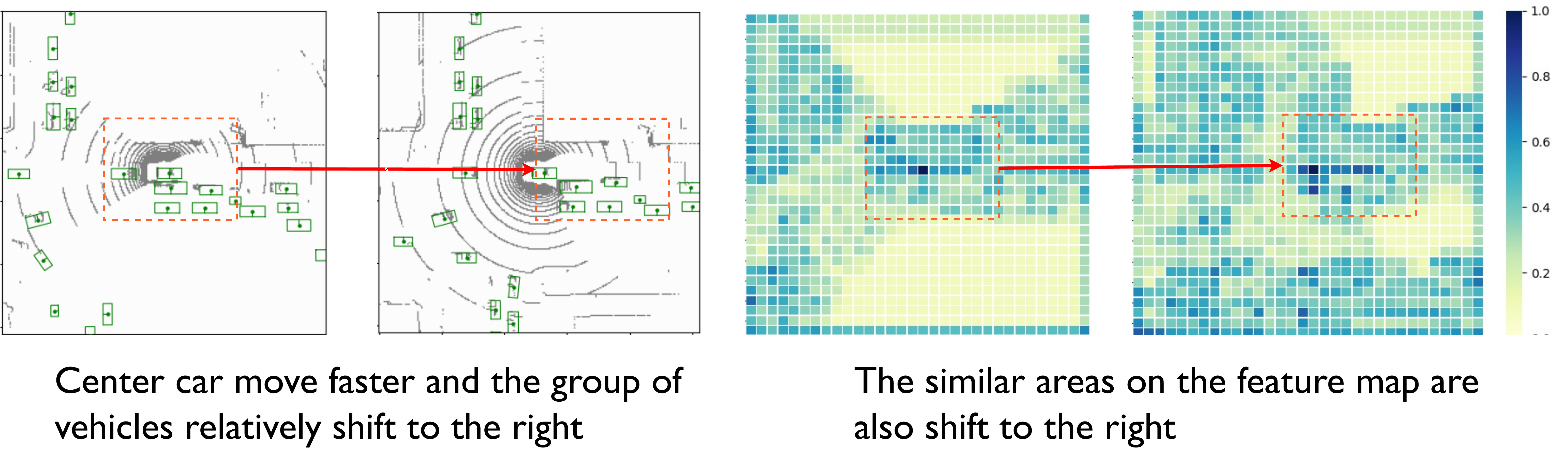}
	\captionsetup{font={small}}
	\caption{Spatial correlation in feature domain. \textcolor{green}{Green} boxes are the ground truth. The heatmap is the feature map $\mathbf{F}\in \mathbb{R}^{H\times W \times C}$ summed over channel dimension. }
	\label{spatial_correlation}
	
\end{figure}

\begin{figure*}[!t]
	\centering
	\begin{subfigure}{0.55\textwidth}
		\centering
		\includegraphics[height=3.8cm]{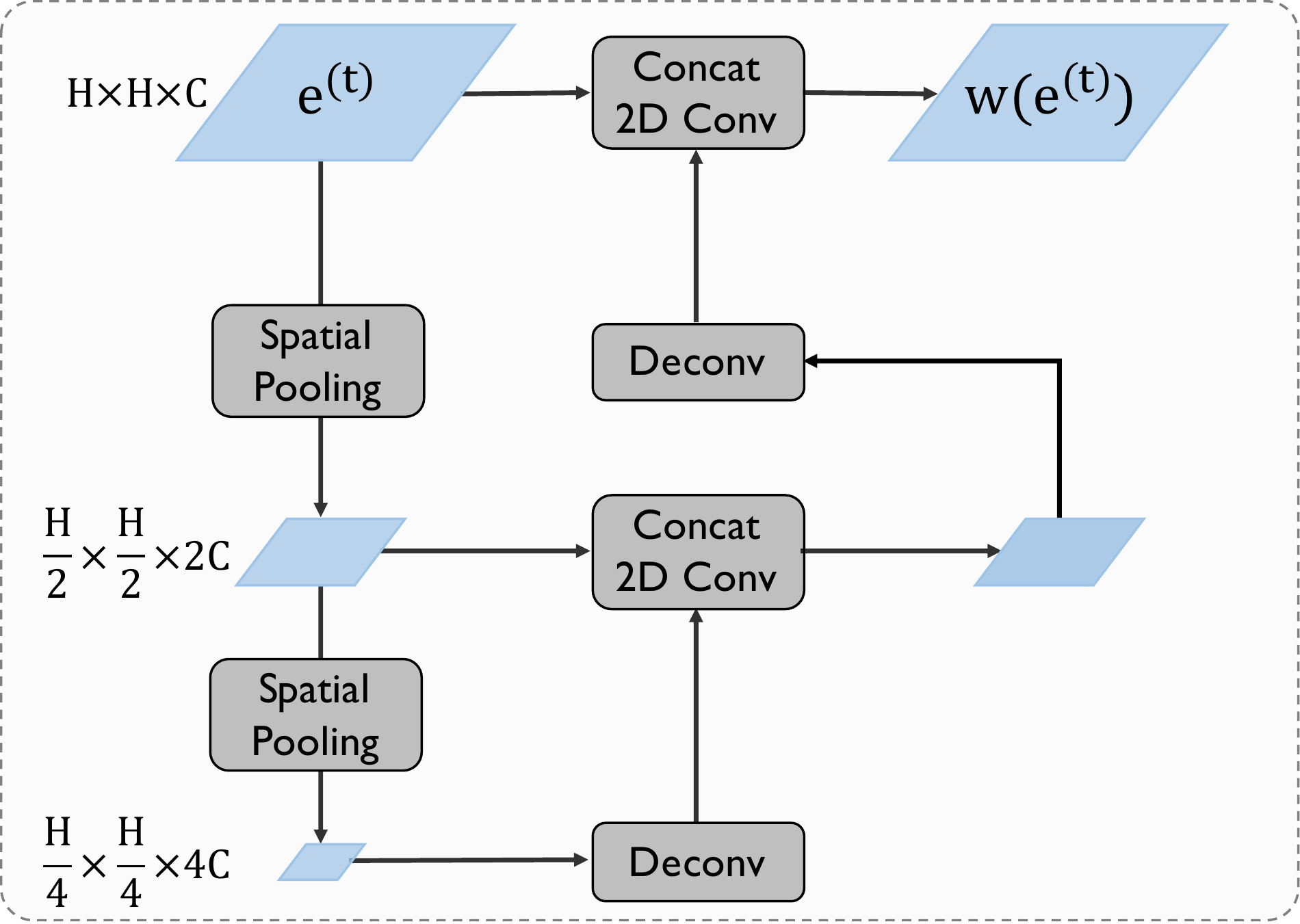}
		\captionsetup{font={small}}
		\caption{Multi-resolution spatial convolution.}
		\label{multiconv}
	\end{subfigure}
    \begin{subfigure}{0.4\textwidth}
		\centering
		\includegraphics[height=3.8cm]{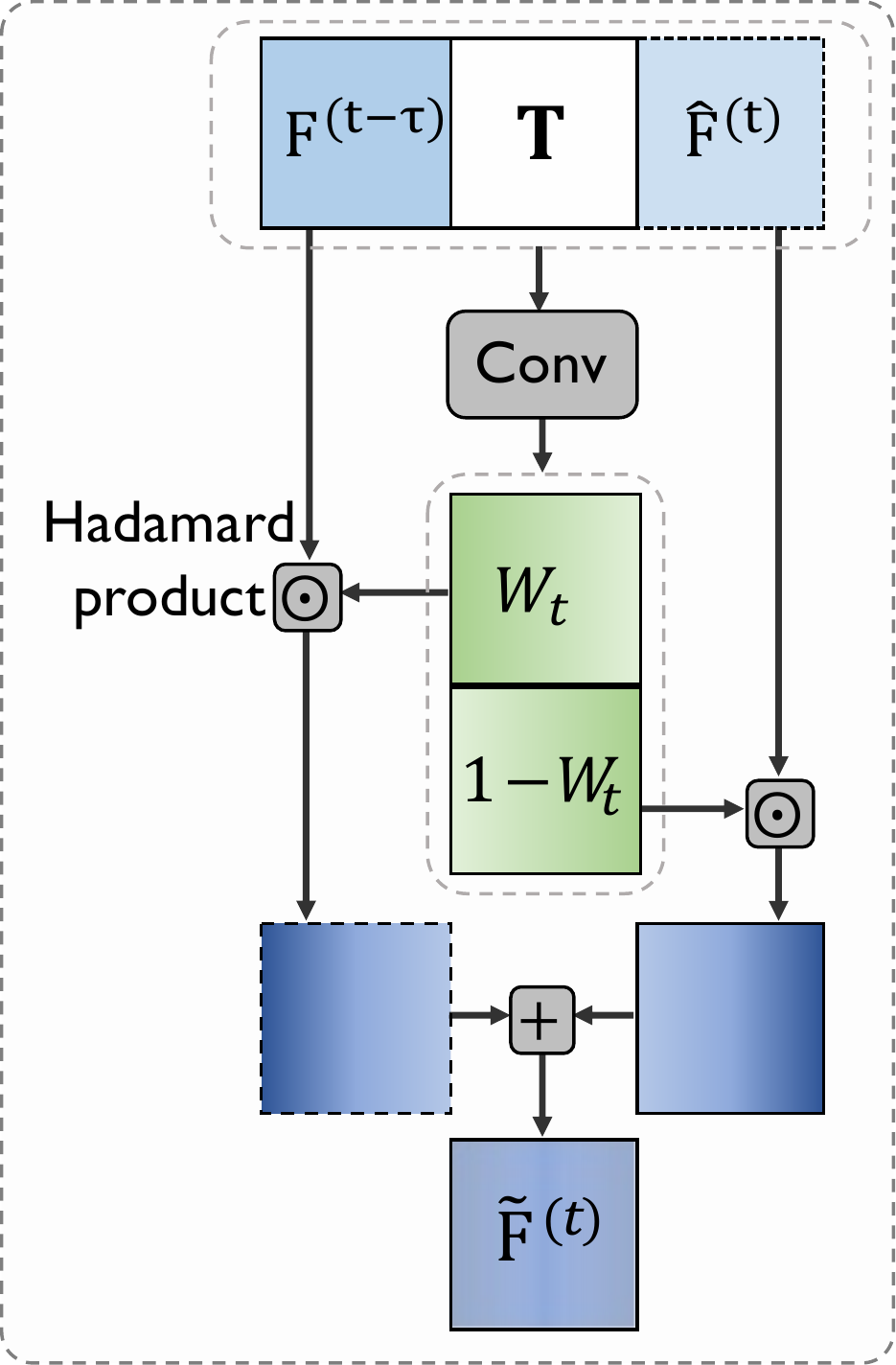}
		\captionsetup{font={small}}
		\caption{Time modulation}
		\label{timemodule}
    \end{subfigure}
	\caption{(a) shows multi resolution spatial convolution in pyramid LSTM. Both $w\left(\cdot\right)$ and $u\left(\cdot\right)$ functions can be represented by this figure. (b) shows time modulation and final estimation feature is $\widetilde{\mathbf{F}}^{\left(t\right)}$. }
\end{figure*}

\noindent\mypar{Feature-attention symbiotic estimation}
Feature-attention symbiotic estimation (FASE) simultaneously estimates the feature and its corresponding collaboration attention by leveraging a novel dual-branch architecture, including feature estimation branch and attention estimation branch. Both branches of the dual-LSTM network share the same input including real-time features perceived by the ego-agent $i$ and $k$ frames of the historical features perceived by its collaborator $j$. 
Each branch is implemented by a pyramid LSTM, which models the series of historical collaborative information and estimates the current state. Pyramid LSTM is specifically designed to capture spatially correlated collaboration features. As shown in Fig.~\ref{spatial_correlation}, when the vehicle group in the red box relatively moves to the right compared with the central vehicle, the same movements occur in similar areas on the feature. The fact shows that the spatial information is significant for our estimation task. We modify the matrix multiplication in LSTM~\cite{lstm} to a multi-scale convolution architecture; see details in Fig.~\ref{multiconv}. The main differences between the proposed pyramid LSTM and the ordinary ones are that LSTM~\cite{lstm} does not specifically consider extracting spatial features; conv-LSTM\cite{convlstm} extracts spatial features at a single scale; while the proposed pyramid LSTM is designed to capture local-to-global features at multiple scales.

The feature estimation branch aims to obtain the most informative features for collaboration at current time. To achieve this, the feature estimation branch should be attention-aware. And the attention estimation branch aims to find the most informative areas for collaboration at current time, besides, it has to suppress the areas with large estimation errors. To achieve this, the attention estimation branch should be feature-aware. To allow the estimation of feature and the corresponding attention be aware of each other, we recurrently leverage both the estimated feature map and collaboration attention from the previous time stamp to be the input of the following time stamp for either branch. 


The entire process is shown in Algorithm~\ref{fase}, where $\tau$ is the latency time, $k$ be the historical frames, $t_0$ be the current time, $\mathbf{W}_j^{\left(t\right)}$ and $\mathbf{F}_j^{\left(t\right)}$ are the collaboration attention and feature from $j$th agent to $i$th agent at time stamp $t$, respectively, $\widetilde{\mathbf{F}}_j^{\left(t\right)}$ and $\widetilde{\mathbf{W}}_j^{\left(t\right)}$ are the estimation of collaboration attention and feature at time stamp $t$, respectively, $\mathbf{e}^{\left(t\right)}$ is the input of pyramid LSTM at time stamp $t$, $\mathbf{h}^{\left(t\right)}_F$, $\mathbf{c}^{\left(t\right)}_F$, $\mathbf{h}^{\left(t\right)}_W$ and $\mathbf{c}^{\left(t\right)}_W$ are the the hidden states and cell states of the pyramid LSTM in each branch, respectively. 

The proposed feature-attention symbiotic estimation has three characteristics: i) the dual-branch structure simultaneously infers the collaboration features and the corresponding collaboration  attention, keeping independent and eliminating the cascaded failure; ii) the estimation networks take the collaboration attention as input so that to focus on more informative areas, promoting more effective estimation; iii) the learnable attention estimation network obtains the information of the feature and gets supervision from attention and fusion feature under ideal collaboration. During the end-to-end optimization, it can not only imitate the weight distribution calculated without latency, but also actively learn to reduce the attention of the spatial position with large noise in features.

\SetKwInput{KwInit}{Initialization}
\SetKwInput{KwPara}{Parameters}
\SetKwInput{KwServer}{Server}

\SetKwProg{ProgServer}{Input:}{:}{}
\SetKwProg{ProgClient}{Function Client Updates}{}{}

\SetKwFunction{Fupdating}{Client Updating}
\SetKwFunction{Fcalculation}{Local anchors Calculation}
\SetKwProg{Fn}{Function}{:}{}
\begin{small}
\begin{algorithm}[t]
\caption{Feature-attention symbiotic estimation}
\label{fase}
\KwPara{Frames of numbers of historical features $k$, frames of latency time $\tau$}
\KwInit{$\mathbf{h}_W^{\left(t_0-\tau-k\right)},~\mathbf{c}_W^{\left(t_0-\tau-k\right)},\mathbf{h}_{F}^{\left(t_0-\tau-k\right)},~\mathbf{c}_F^{\left(t_0-\tau-k\right)}$.}
\ProgServer{$\mathbf{F}_i^{\left(t_0\right)},\left\{\mathbf{F}_j^{\left(t^\prime\right)}\right\}_{t^\prime = t-\tau-k+1,...,t-\tau}$}{
    \For{$t=t_0-\tau-k+1,t_0-\tau-k+2,...,t_0-\tau-1$}{

        $\mathbf{W}^{\left(t\right)}=f_{\rm attention}(\mathbf{F}_i^{t_0},\mathbf{F}_j^{t}),\ \mathbf{e}^{\left(t\right)} = \mathbf{F}_j^{\left(t\right)}\left|\right.\mathbf{W}_j^{\left(t\right)}$
        
        
        $\left(\mathbf{h}_F^{\left(t\right)},\mathbf{c}_F^{\left(t\right)}\right) = p_F\left(\mathbf{e}^{\left(t\right)},\left(\mathbf{h}_F^{\left(t-1\right)},\mathbf{c}_F^{\left(t-1\right)}\right)\right)$
        
        $\left(\mathbf{h}_W^{\left(t\right)},\mathbf{c}_W^{\left(t\right)}\right) = 
        p_W\left(\mathbf{e}_W^{\left(t\right)},\left(\mathbf{h}_W^{\left(t-1\right)},\mathbf{c}_W^{\left(t-1\right)}\right)\right)$
    }
    \For{$t=t_0-\tau+1,t_0-\tau+2,...,t_0-1$}{
        $\widetilde{\mathbf{F}}_j^{\left(t\right)} = d_{\rm F}(\mathbf{h}_F^{\left(t-1\right)})$
        \hfill $\leftarrow$ \textbf{Estimation of the feature at $t$}
        
        $\widetilde{\mathbf{W}}_j^{\left(t\right)} = d_{\rm W}(\mathbf{h}_W^{\left(t-1\right)})$
        \hfill $\leftarrow$ \textbf{Estimation of the attention at $t$}
        
        $\textbf{e}^{\left(t\right)} = \widetilde{\mathbf{F}}_j^{\left(t\right)}\left|\right.\widetilde{\mathbf{W}}_j^{\left(t\right)}$
        
        $\left(\mathbf{h}_F^{\left(t\right)},\mathbf{c}_F^{\left(t\right)}\right) = p_F\left(\mathbf{e}^{\left(t\right)},\left(\mathbf{h}_F^{\left(t-1\right)},\mathbf{c}_F^{\left(t-1\right)}\right)\right)$
        
        $\left(\mathbf{h}_W^{\left(t\right)},\mathbf{c}_W^{\left(t\right)}\right) = 
        p_W\left(\mathbf{e}_W^{\left(t\right)},\left(\mathbf{h}_W^{\left(t-1\right)},\mathbf{c}_W^{\left(t-1\right)}\right)\right)$

    }
$\widetilde{\mathbf{F}}_j^{\left(t_0\right)} = d_{\rm F}(\mathbf{h}_F^{\left(t_0-1\right)}),\ \widetilde{\mathbf{W}}_j^{\left(t_0\right)} = d_{\rm W}(\mathbf{h}_W^{\left(t_0-1\right)})$

}
\end{algorithm}
\end{small}

\noindent\mypar{Time modulation}
Although FASE achieves the basic functionality of $c\left(\cdot\right)$, 
we find, when latency is low, the performance degradation caused by latency is relatively minor than the estimation noise led by FASE. To handle this, we propose time modulation, it attentively fuses the raw (working well at low latency) and estimated (working well at high latency) features conditioned on the latency time, generating more comprehensive and reliable estimation. 

Let $\mathbf{M}_F^{\left(t\right)},\mathbf{M}_W^{\left(t\right)} \in \mathbb{R}^{H \times W}$ be a confidence matrix to reflect the estimation uncertainty level of each spatial region, $\mathbf{T}_F \in \mathbb{R}^{H \times W \times C}$ and $\mathbf{T}_W \in \mathbb{R}^{H \times W}$ be the latency tensor obtained from the expansion of latency time $\tau \in \mathbb{R}$ with the same shape as $\widetilde{\mathbf{F}}^{\left(t\right)}$ and $\widetilde{\mathbf{W}}^{\left(t\right)}$, respectively. Time modulation works as 
\begin{subequations}
	\small
\begin{equation}
    \mathbf{M}_F^{\left(t\right)} = m_F\left(\widetilde{\mathbf{F}}^{\left(t\right)}\left|\right.\mathbf{F}^{\left(t-\tau\right)}\left|\right.\mathbf{T}_W\right), \mathbf{M}_W^{\left(t\right)} = m_W\left(\widetilde{\mathbf{W}}^{\left(t\right)}\left|\right.\mathbf{W}^{\left(t-\tau\right)}\left|\right.\mathbf{T}_F\right),
    \label{timeweight}
\end{equation}
\begin{equation}
    \dot{\widetilde{\mathbf{F}}}^{\left(t\right)}=\widetilde{\mathbf{F}}^{\left(t\right)} \odot \mathbf{M}_F^{\left(t\right)} + \mathbf{F}^{\left(t-\tau\right)} \odot (\mathbf{1}-\mathbf{M}_F^{\left(t\right)}), \dot{\widetilde{\mathbf{W}}}^{\left(t\right)}=\widetilde{\mathbf{W}}^{\left(t\right)} \odot \mathbf{M}_W^{\left(t\right)} + \mathbf{W}^{\left(t-\tau\right)} \odot (\mathbf{1}-\mathbf{M}_W^{\left(t\right)}),
    \label{timesum}
\end{equation}
\end{subequations}
where $m_F\left(\cdot\right)$ and $m_W\left(\cdot\right)$ are both the lightweight convolution neural network with a sigmoid activation function, $\mathbf{1} \in \mathbb{R}^{H \times W}$ is a matrix in which all elements are 1. Step~\eqref{timeweight} respectively obtains the confidence level of the feature estimation at each spatial region by leveraging the concatenation of the estimated collaborative feature/attention by FASE, the latest asynchronous feature/attention, and the latency tensor. According to the confidence matrix, Step~\eqref{timesum} respectively combines the estimated feature/attention and the latest asynchronous feature/attention. We expect when the latency is high, the confidence matrix would have higher weights and the estimated feature/attention would contribute more to the final estimation; also see the process in Fig.~\ref{timemodule}. 
\subsection{Loss function}
\label{sec:loss}
Let $\mathbf{Y}_i^{(t)}$ be the ground truth of final perception output of the $i$th agent at time stamp $t$, $\mathbf{H}_i^{(t)}$ be the ground truth feature of the $i$th agent at time stamp $t$ after aggregating real-time collaboration information, $\mathbf{F}_i^{(t)}$ be the ground truth feature map of the $i$th agent at time stamp $t$, and $\mathbf{W}_{j\rightarrow i}^{\left(t\right)}$ be the ground truth collaboration attention from the $j$th agent to the $i$th agent at time stamp $t$. We consider minimizing the following objective to optimize the overall latency-aware collaborative perception system:

\begin{align*}
	\small
    \mathcal{L}=&\lambda_o\ell_{\rm output} \left(\mathbf{Y}_i^{(t)},  \widetilde{\mathbf{Y}}_i^{(t)} \right) + \lambda_f\ell_{\rm fusion} \left( \mathbf{H}_i^{(t)},  \widetilde{\mathbf{H}}_i^{(t)} \right)+\\
     &\lambda_f\ell_{\rm feature} \left(\mathbf{F}_i^{(t)},\widetilde{\mathbf{F}}_i^{(t)} \right)+
     \lambda_w\ell_{\rm weight} \left(\mathbf{W}_{j\rightarrow i}^{\left(t\right)},\widetilde{\mathbf{W}}_{j\rightarrow i}^{\left(t\right)}\right),
\end{align*}
where $\lambda$ denotes the weight of each item,  $\ell_{\rm output} (\cdot)$ is the ultimate perception loss, $\ell_{\rm fusion} (\cdot), \ell_{\rm feature} (\cdot), \ell_{\rm weight} (\cdot)$ are losses for fused feature, intermediate estimated feature and estimated collaborative attention, respectively. The first term supervises the perception output and the second term supervises the estimated fusion feature.  The third and fourth terms provide more supervision on intermediate feature maps and collaborative attentions to promote faster convergence.

\section{Experiments}

\begin{figure}[!t]
	\centering
	\includegraphics[height=3.8cm]{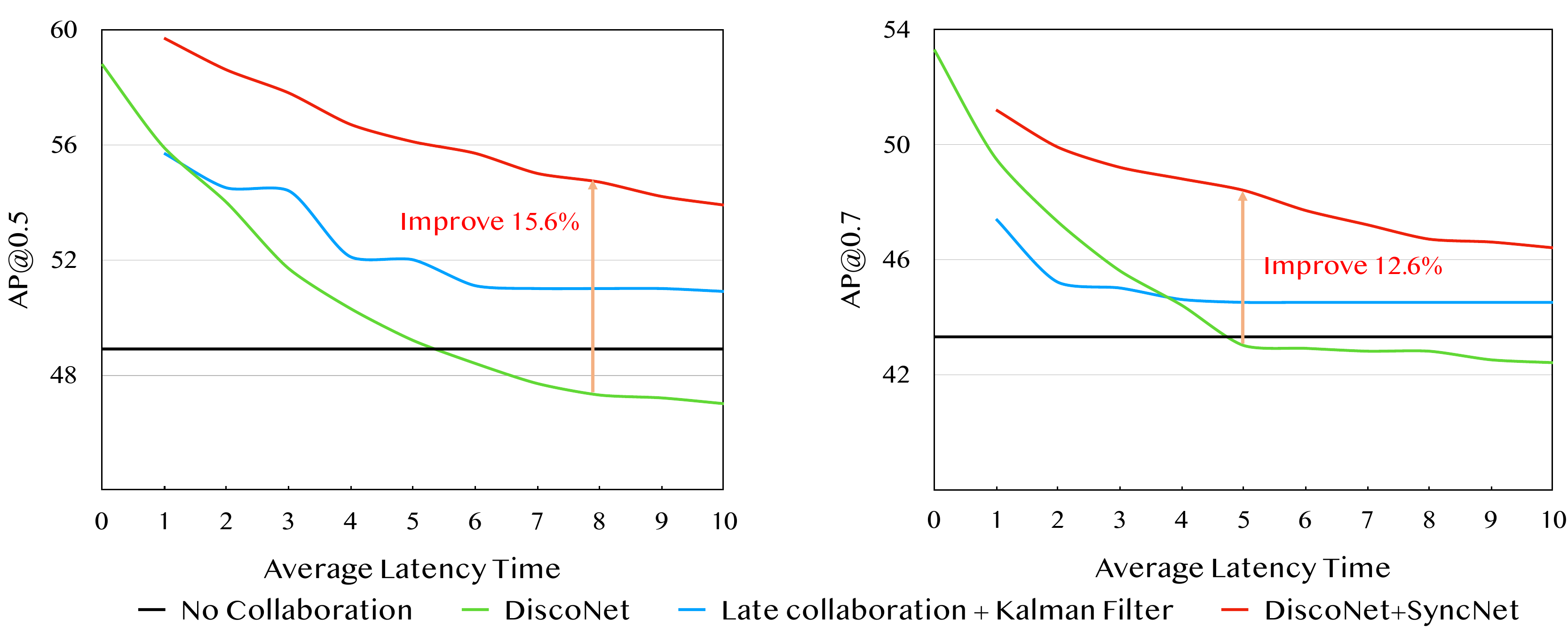}
	\captionsetup{font={small}}
	\caption{Comparison of the performance of No Collaboration, DiscoNet\cite{disconet}, late collaboration with Kalman Filter, DiscoNet + SyncNet in latency of 1-10 frames.}
	\label{fig:comparison}

\end{figure}

\begin{figure*}[!t]
	\centering

	\begin{subfigure}{0.4\textwidth}
		\centering
		\includegraphics[height=3.6cm]{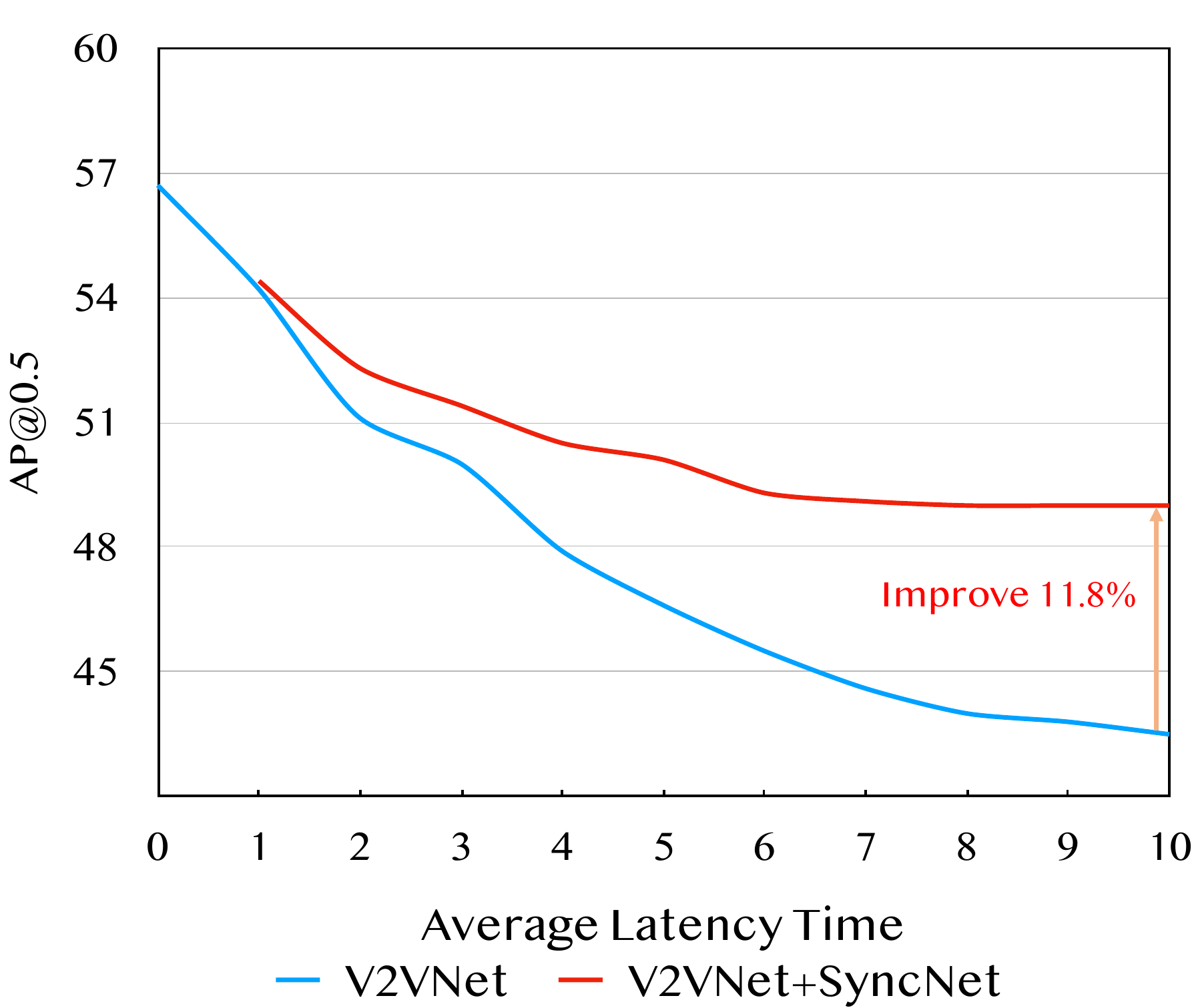}
		\label{v2v}
		\captionsetup{font={small}}
		\caption{V2VNet with/out SyncNet}
	\end{subfigure}
    \begin{subfigure}{0.4\textwidth}
		\centering
   		\includegraphics[height=3.6cm]{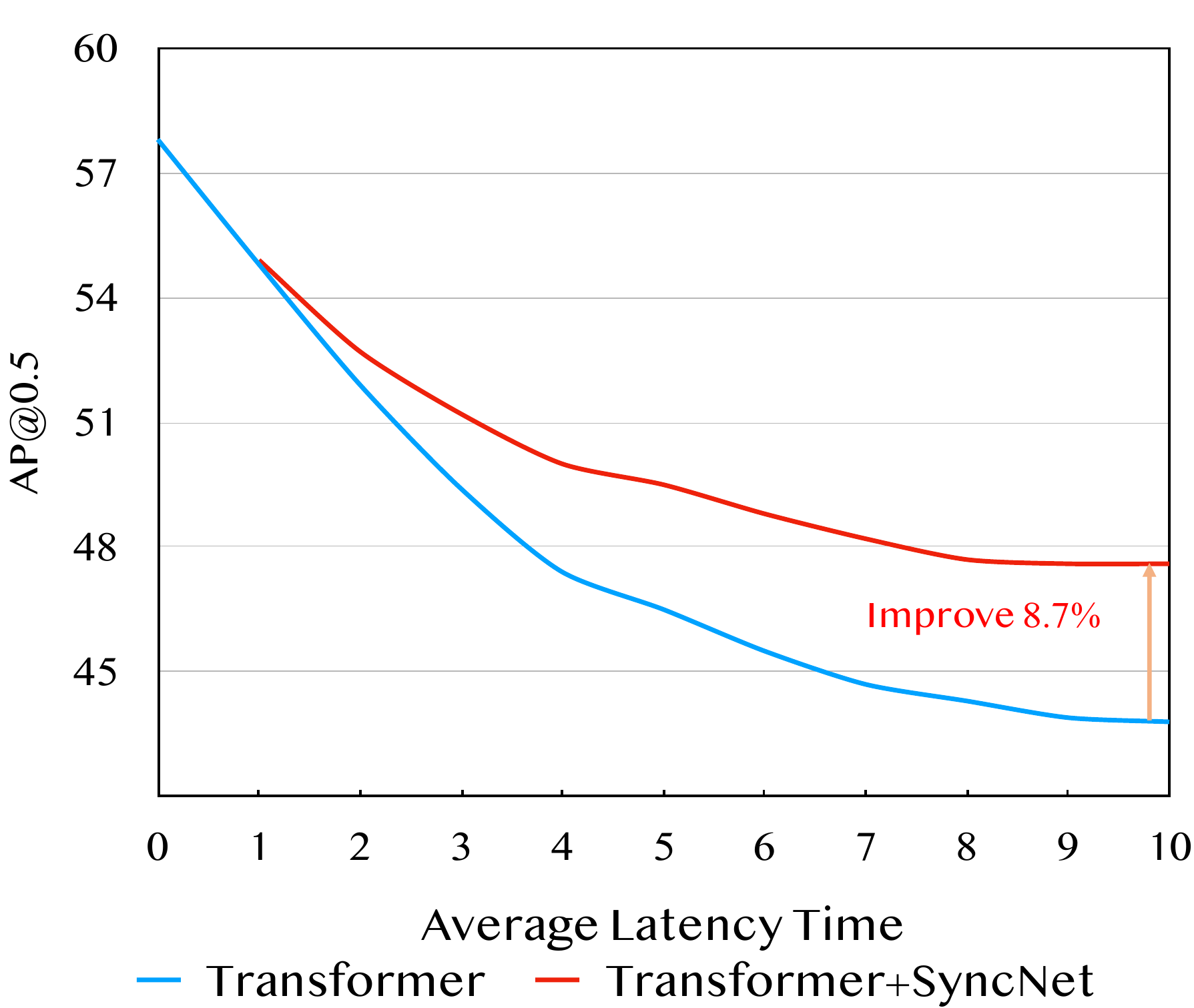}

		\label{transformer_module}
		\captionsetup{font={small}}
		\caption{Transformer with/out SyncNet}
    \end{subfigure}
	\captionsetup{font={small}}
	\caption{SyncNet integrated with different fusion frameworks in AP 0.5.}
	\label{transformer}

\end{figure*}

\begin{figure}[!t]
\begin{minipage}[b]{0.28\textwidth}
    \centering
    \includegraphics[width=\textwidth]{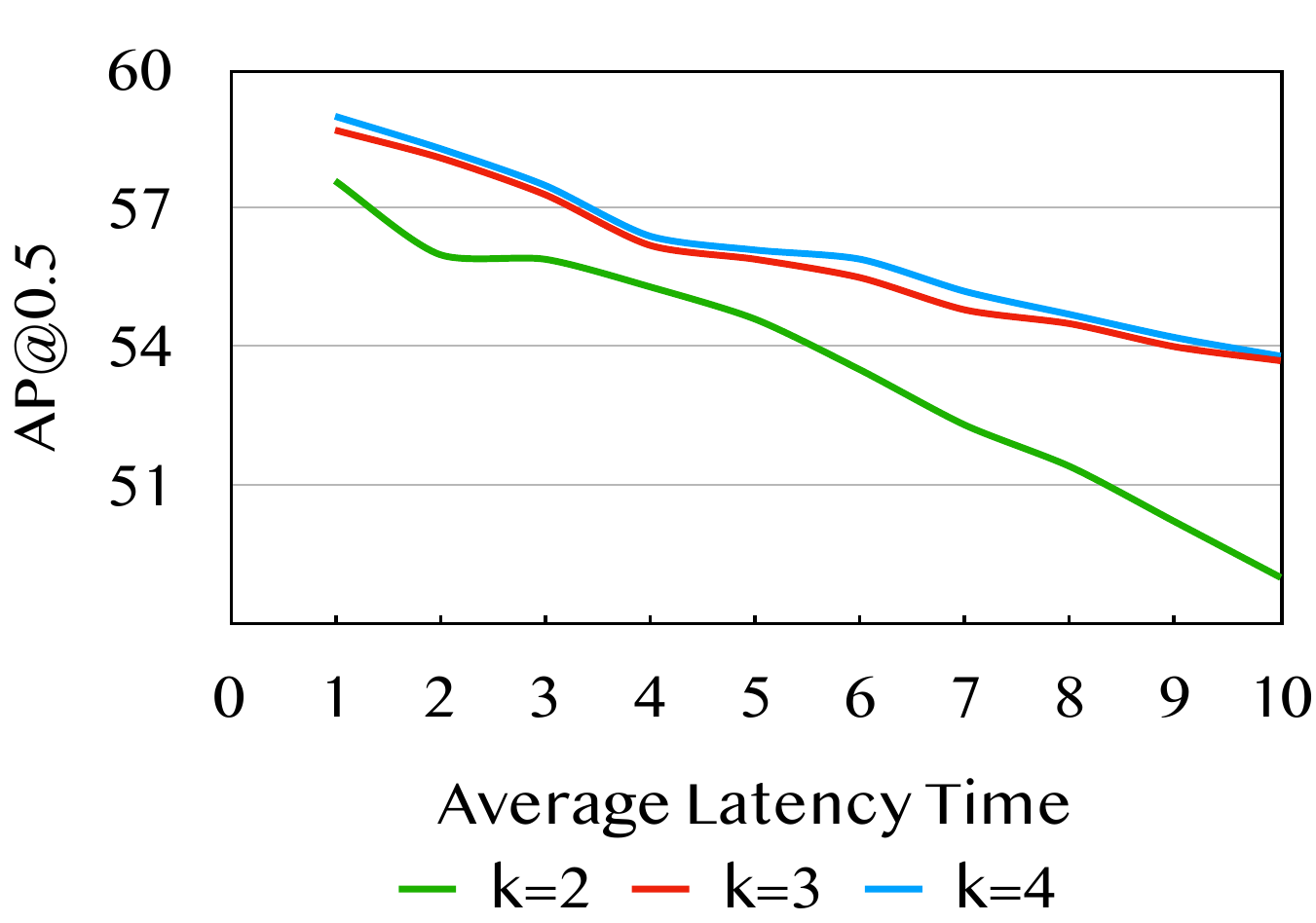}
	\captionsetup{font={small}}
    \caption{Ablation of historical frames $k$.}
    \label{k234}
\end{minipage}
\begin{minipage}[b]{0.63\linewidth}
\centering
\setlength\tabcolsep{1.5pt}
\scriptsize
    \centering
     \begin{tabular}{c | c|c|c||l|l}
\toprule
Method  & Compensation  &   LSTM & TM       &    $\tau = 1$ &  $\tau = 5$\\
\hline
A &  Vanilla & Single-Scale &\checkmark& 56.0/47.8 & 54.6/44.2\\
B &  Vanilla &  Pyramid & & 56.1/48.3 & 54.6/44.2\\
C &  Vanilla &  Pyramid     &\checkmark& 60.1/48.9    & 55.6/45.1\\
D &   FASE&   Single-Scale  &\checkmark& 59.1/50.8     &   55.3/48.8\\
E &  FASE &  Pyramid     &&   55.7/48.4    &   54.3/47.6\\
F &  FASE &  Pyramid     &\checkmark&   59.7/51.2    &   56.1/48.8\\
\toprule
\end{tabular}
\captionsetup{font={small}}
\captionof{table}{ Ablation of SyncNet\\ in AP@0.5/0.7.}
\label{maintable}
\end{minipage}

\end{figure}

\begin{figure}[!t]

	\centering
	\includegraphics[height=3.8cm]{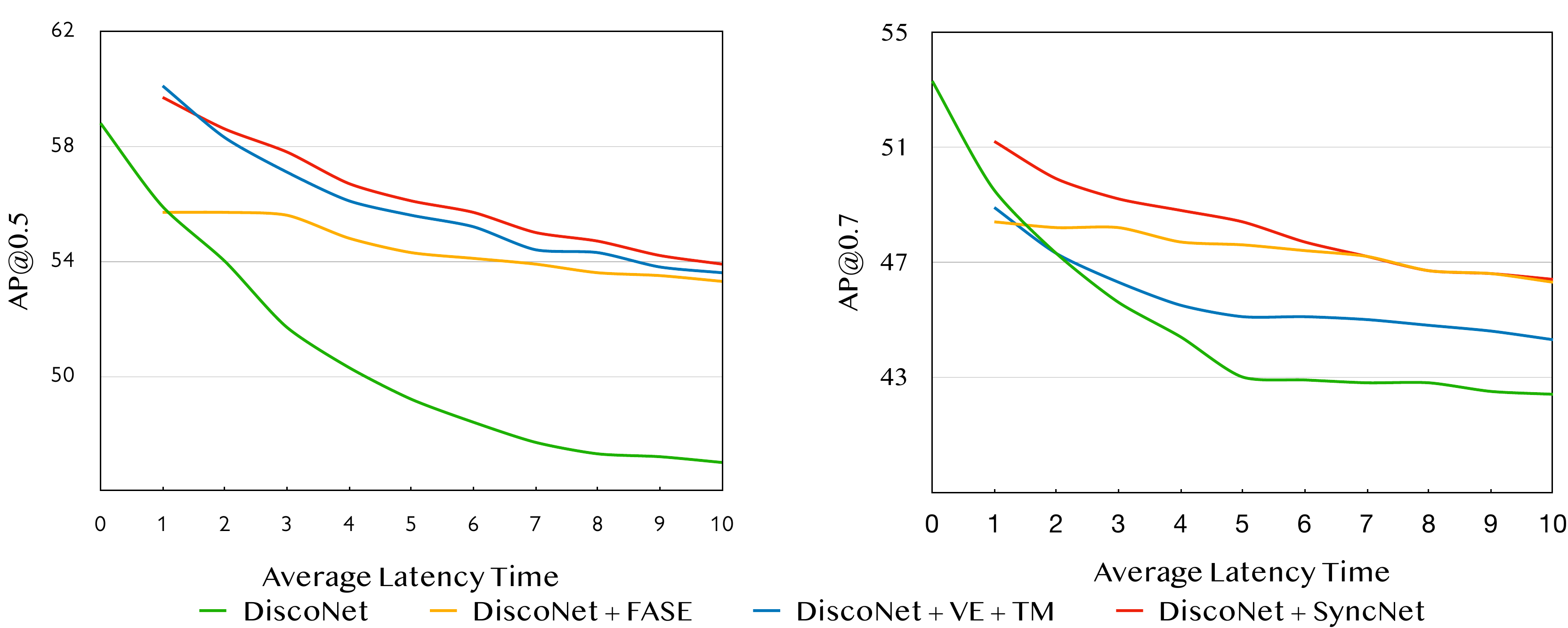}
	\captionsetup{font={small}}
	\caption{Ablation study: compare \textit{DiscoNet}, \textit{DiscoNet+FASE}, \textit{DiscoNet+Vanilla Estimation+Time Modulation}, \textit{DiscoNet+SyncNet}. FASE can achieve improvement for AP@0.7 and time modulation can achieve improvement at low latency.}
	\label{tm_com}

\end{figure}

\begin{figure}[t]
	\centering
	\begin{subfigure}{0.24\textwidth}
		\centering
		\includegraphics[height=5.5cm]{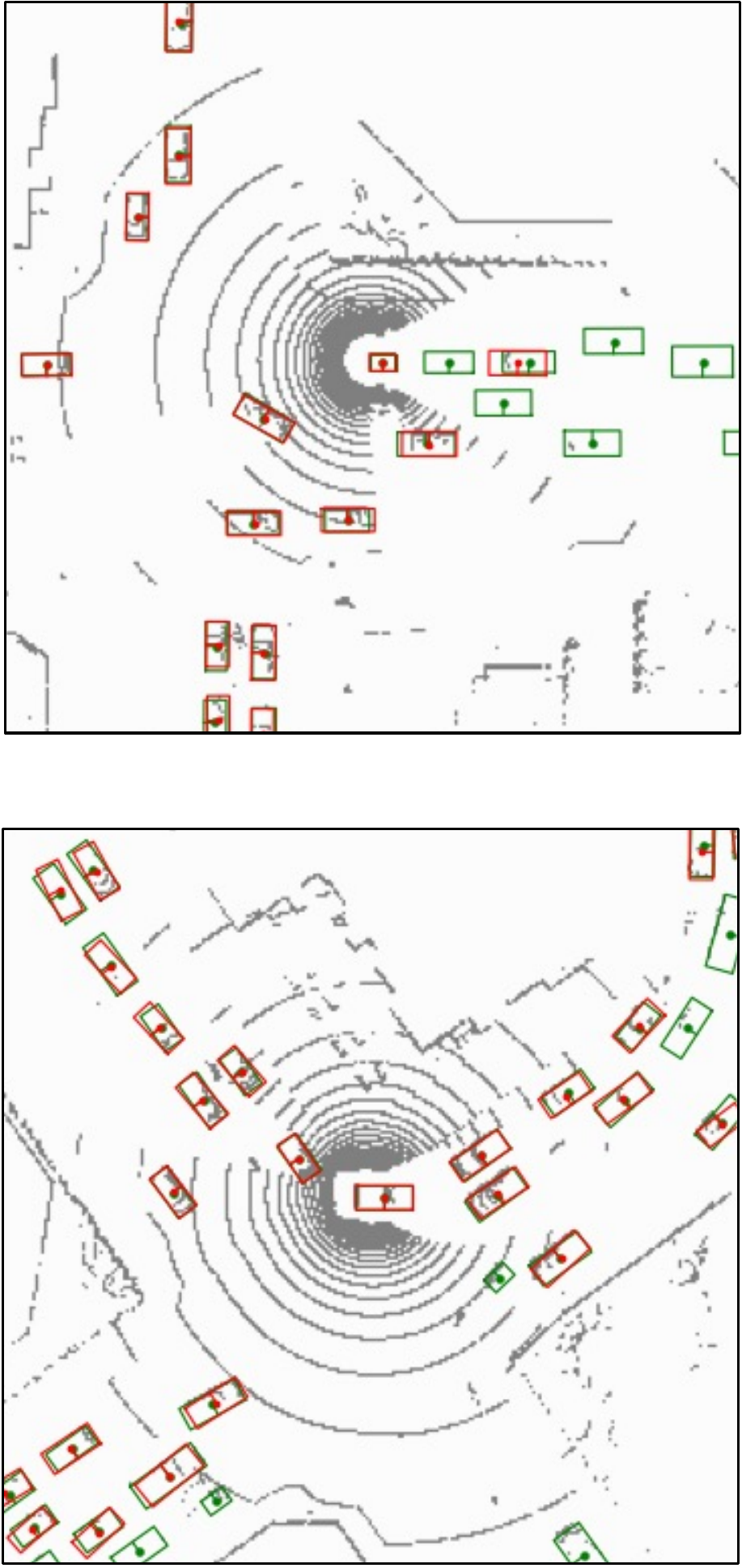}
		\label{fig:0latency}
		\captionsetup{font={small}}
		\caption{\textit{DiscoNet}\\(without latency)}
	\end{subfigure}
    \begin{subfigure}{0.24\textwidth}
		\centering
   		\includegraphics[height=5.5cm]{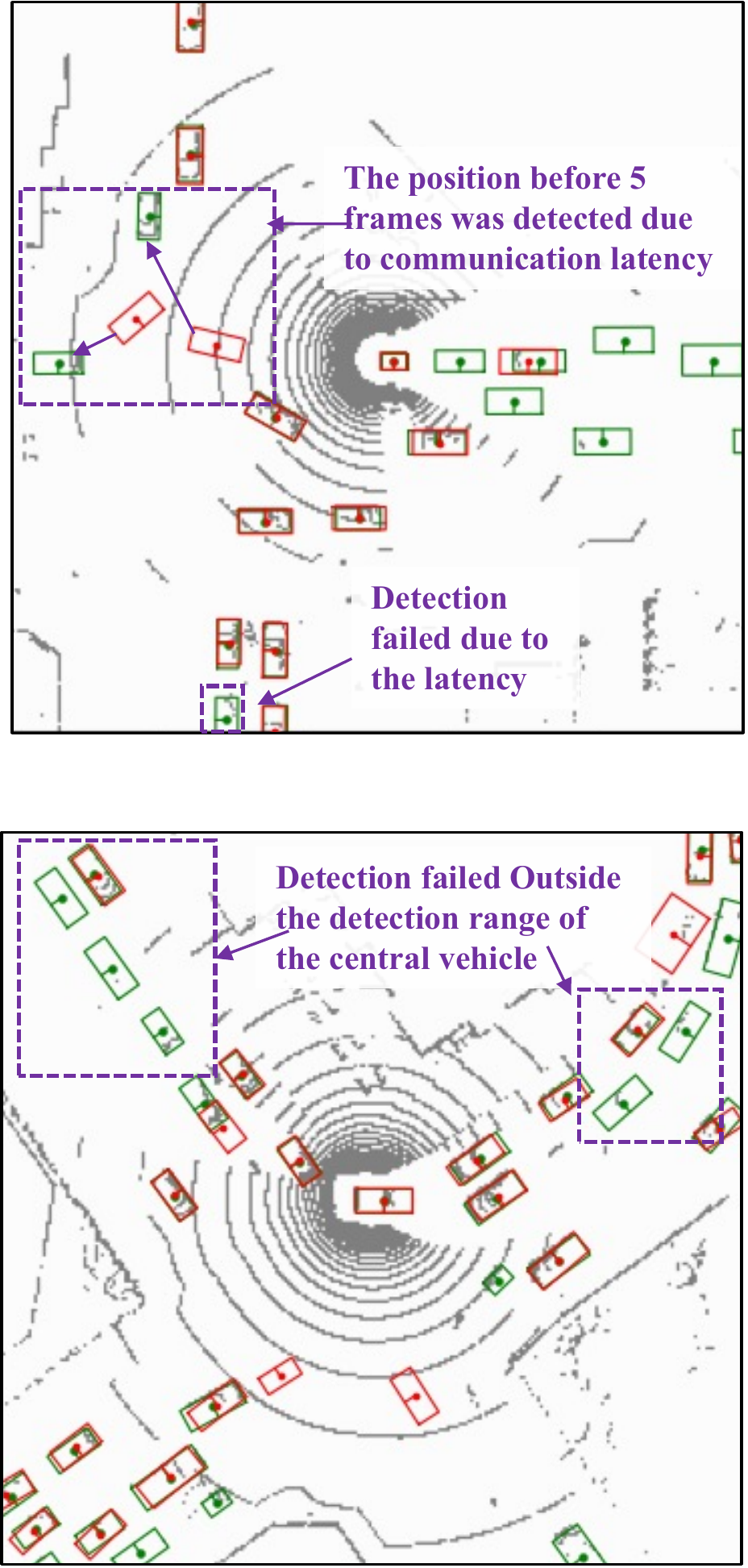}
		\label{fig:5latency}
		\captionsetup{font={small}}
		\caption{\textit{DiscoNet}\\(latency)}
    \end{subfigure}
    \begin{subfigure}{0.24\textwidth}
		\centering
   		 \includegraphics[height=5.5cm]{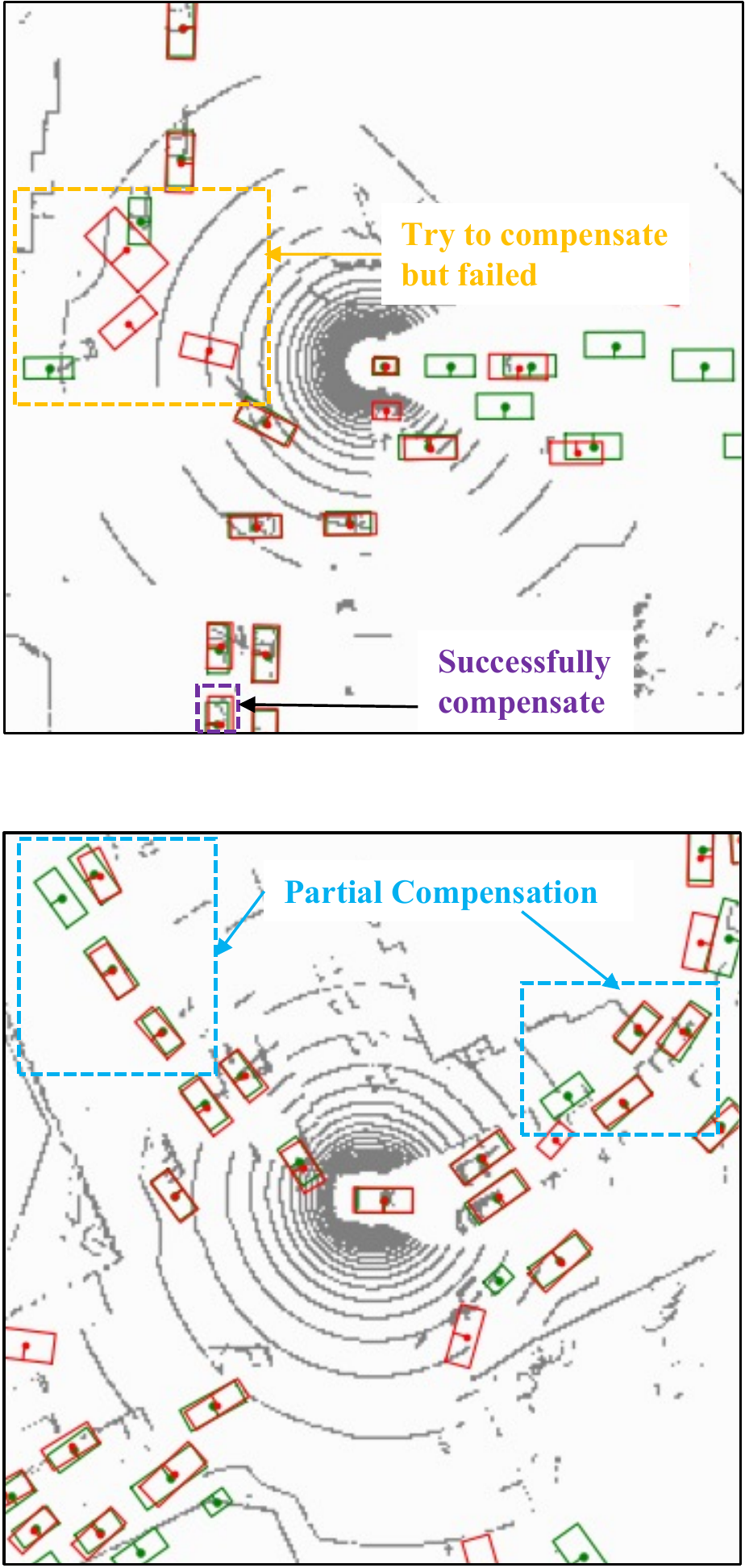}
		\label{fig:stpnlstm}
		\captionsetup{font={small}}
		\caption{\textit{DiscoNet+VE}\\(latency)}
    \end{subfigure}
    \begin{subfigure}{0.24\textwidth}
		\centering
   		 \includegraphics[height=5.5cm]{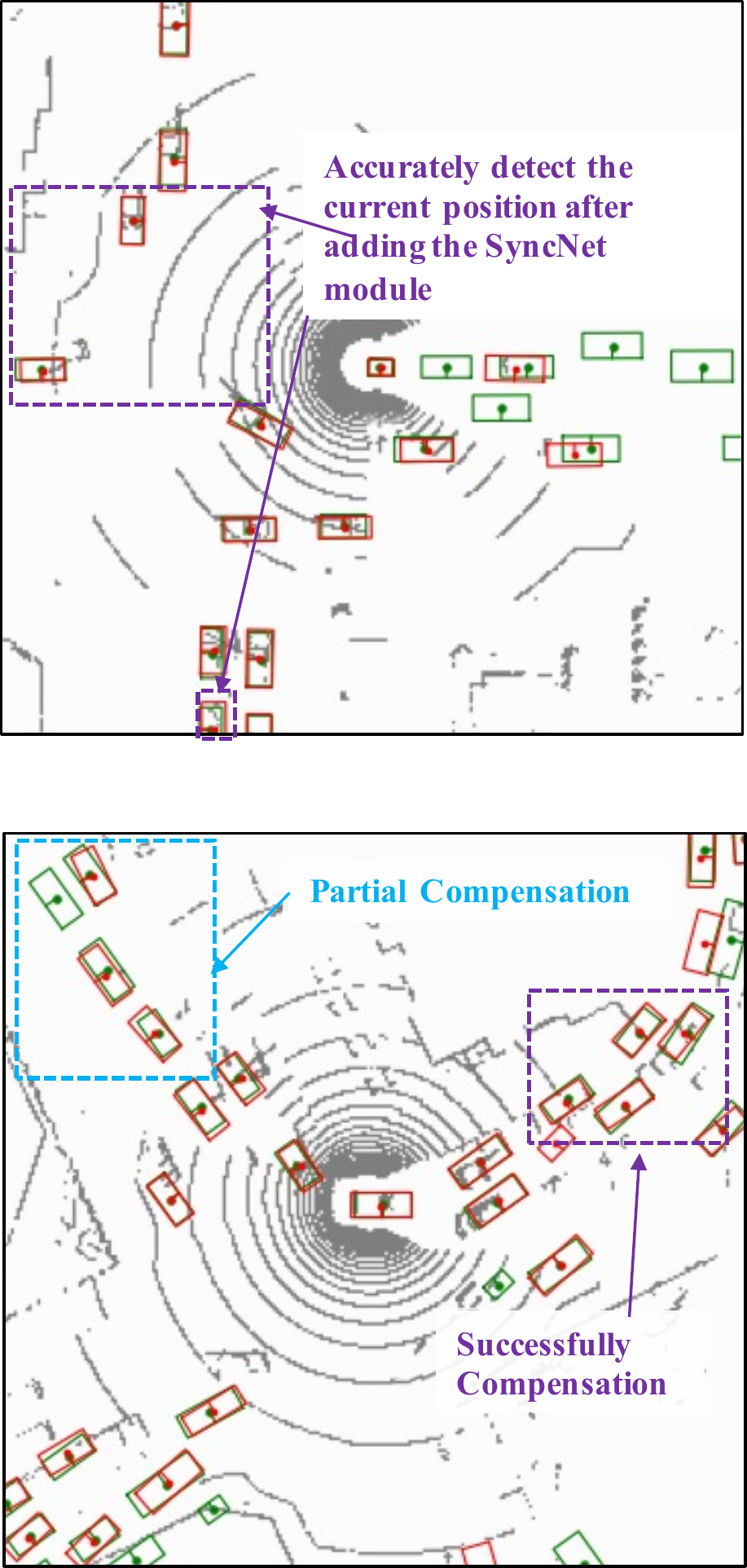}
		\label{fig:syncnet}
		\captionsetup{font={small}}
		\caption{\textit{DiscoNet+Sync\\Net}(latency)}
    	\end{subfigure}
	\captionsetup{font={small}}
	\caption{FASE architecture qualitatively improves the performance under communication latency. (a) shows the detection results of DiscoNet~\cite{disconet} without communication latency. (b) (c) (d) show results in the case of the latency of an average of 5 frames.}

	\label{visual_car}
\end{figure}
\begin{figure}[t]
	\centering
	\begin{subfigure}{0.232\textwidth}
		\centering
		\includegraphics[height=2.47cm]{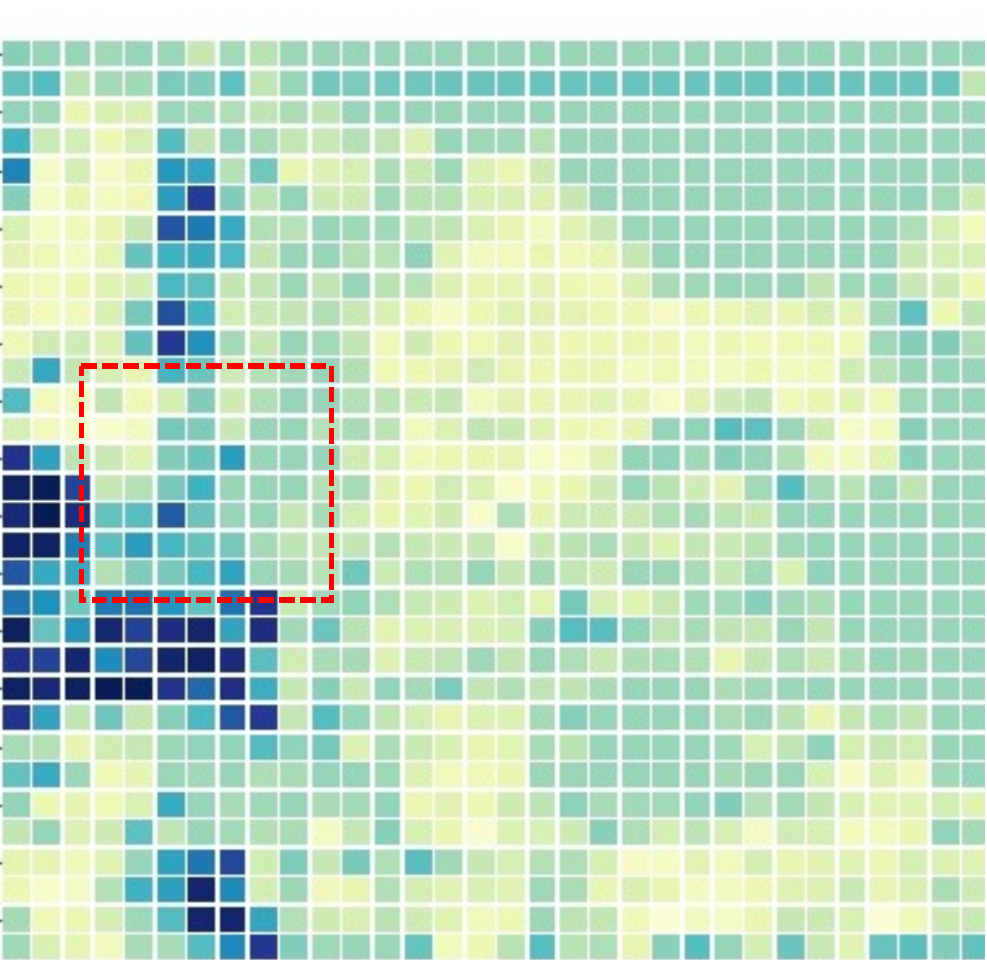}
		\label{fig:0latency}
		\captionsetup{font={small}}
		\caption{\textit{DiscoNet}\\(without latency)}
	\end{subfigure}
    \begin{subfigure}{0.229\textwidth}
		\centering
   		\includegraphics[height=2.44cm]{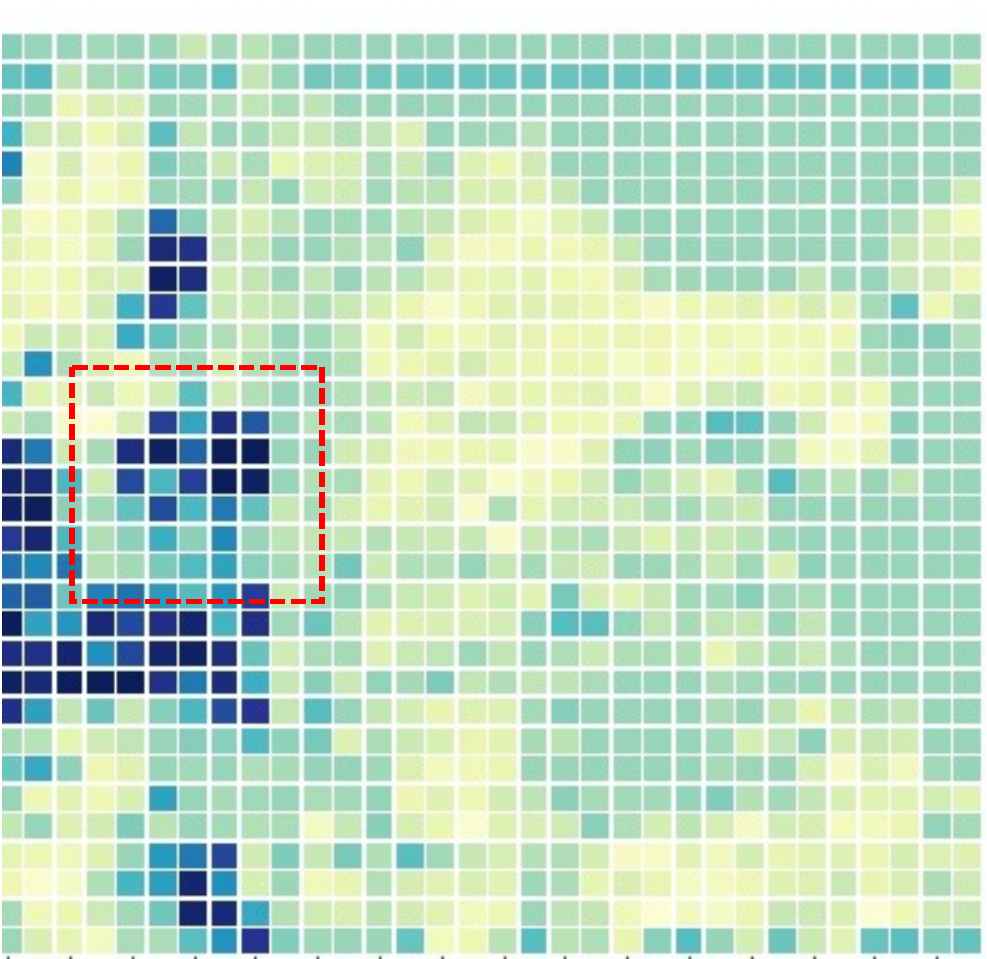}
		\label{fig:5latency}
		\captionsetup{font={small}}
		\caption{\textit{DiscoNet}\\(latency)}
    \end{subfigure}
    \begin{subfigure}{0.2325\textwidth}
		\centering
   		 \includegraphics[height=2.455cm]{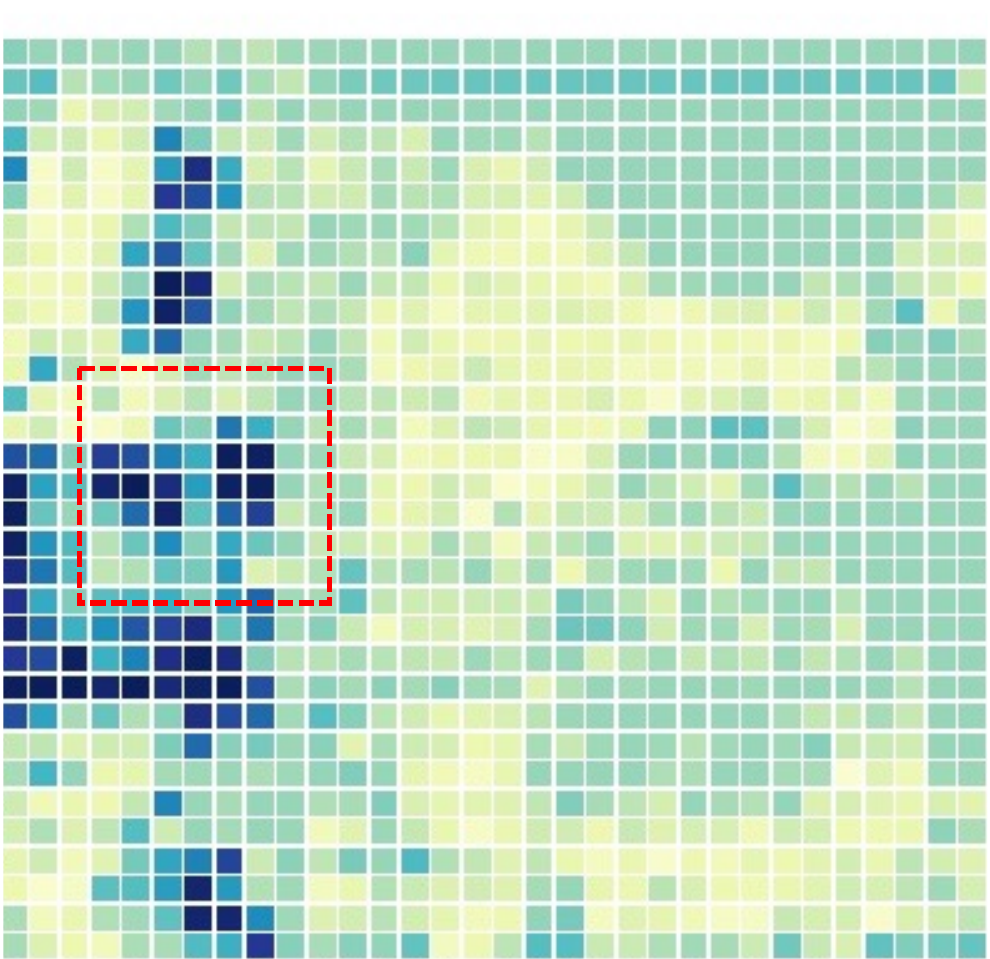}
		\label{fig:stpnlstm}
		\captionsetup{font={small}}
		\caption{\textit{DiscoNet+VE}\\(latency)}
    \end{subfigure}
    \begin{subfigure}{0.28\textwidth}
   		 \includegraphics[height=2.44cm]{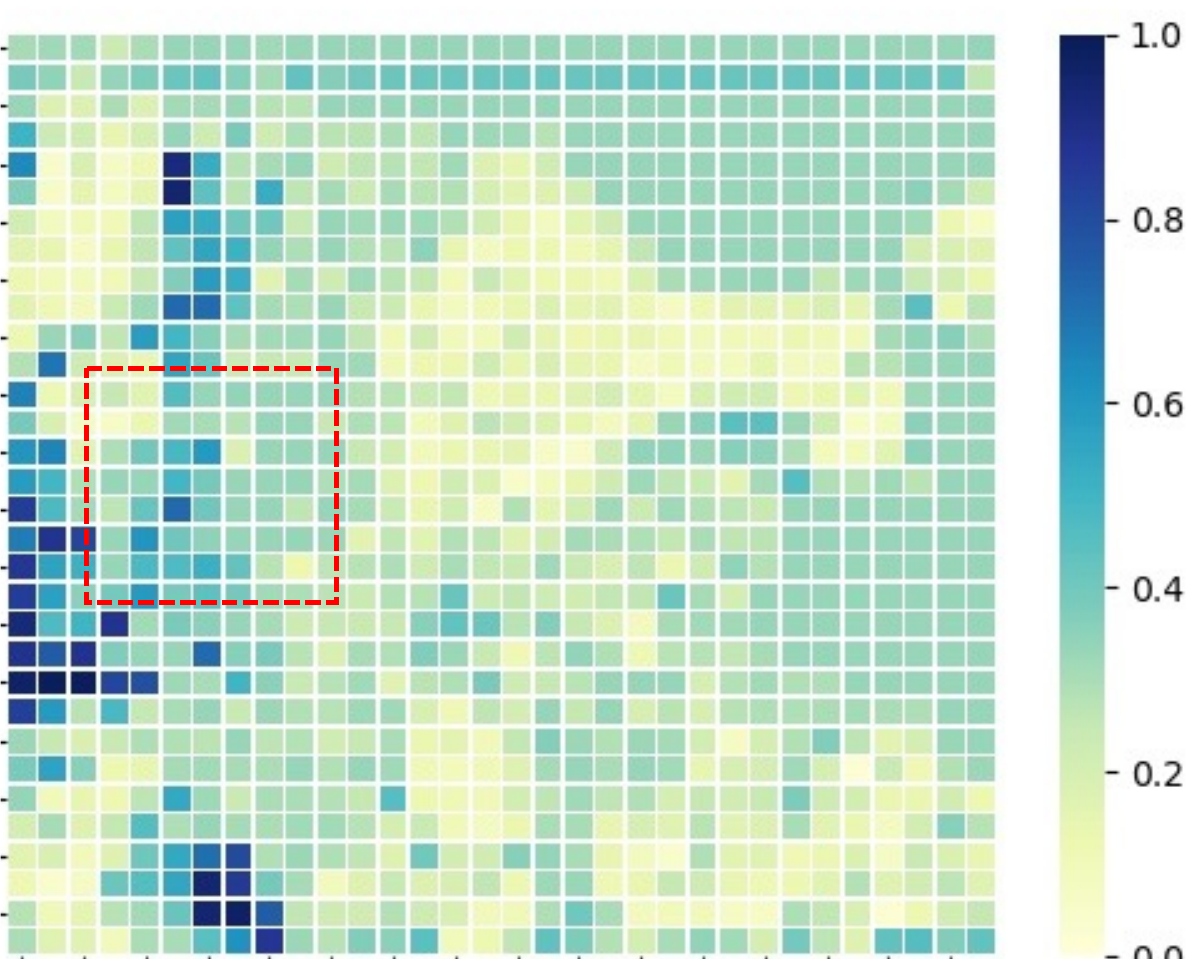}
		\label{fig:syncnet}
		\captionsetup{font={small}}
		\caption{\textit{DiscoNet+SyncNet}\\(latency)}
    	\end{subfigure}
	\captionsetup{font={small}}
    \caption{Collaboration attention of a feature sent by a neighbor agent in the scenario of the first row in Fig.\ref{visual_car}. (b) (c) (d) show results in the case of the latency of an average of 5 frames. Comparing (b), (c) and (d) we see that SyncNet obtains a feature which is closer to (a), and actively reduces the weight of the noisy position in the red box.}
	\label{weightstudy}

\end{figure}

\subsection{Multi-agent 3D object detection dataset}
We validate our SyncNet on LIDAR-based 3D object detection task\cite{pointcloud,pcsurvey} with a multi-agent dataset, V2X-Sim\cite{v2xsim}. V2X-Sim is built with the co-simulation of SUMO\cite{sumo} and CARLA\cite{carla}. V2X-Sim includes 80 scenes in the training set and 11 scenes in the test set. Each sample contains 2.67 agents on average and includes 3D point clouds input and 3D bounding box annotations. The 3D point clouds are generated by a LIDAR with 32 channels and 70m max range, 20Hz rotation frequency, and 5Hz recorded frequency. To simulate collaborative perception under a latency scenario, we load data in asynchronous time stamps, and the latency time $\tau$ is randomly generated from an exponential distribution.

\subsection{Implementation details}
\textbf{Experimental setting.} We crop the point clouds which locate in the region of $[-32m,32m]\times[-32m,32m]\times[0,5m]$ defined in the ego-vehicle Cartesian coordinate system. We set the size of each voxel as $0.25m\times0.25m\times0.4m$. After crop and voxelization, we get a Bird's-Eyes view map with dimension $256\times256\times13$. The encoded features to be transmitted has a dimension of $32\times32\times256$. The latency between two agents can be a fixed or random number generated by exponential distribution rounded to an integer. We train our model using NVIDIA RTX 3090 GPU with Pytorch. The evaluation metric is the Average Precision(AP) metric at Intersection-over-Union threshold of 0.5 and 0.7.

\textbf{Baselines.} 
Our proposed latency-aware collaborative perception system adopts one of the state-of-the-art collaborative perception frameworks, DiscoNet~\cite{disconet}, and leverages the proposed SyncNet as the latency compensation module to handle various latency settings. To validate our latency-aware collaborative perception system, \textit{DiscoNet+SyncNet}, we compare with three baselines: i) single-agent perception system, \textit{No Collaboration}; ii) latency-unaware collaborative perception,~\textit{DiscoNet}~\cite{disconet}; iii) naive latency-aware late-fusion-based collaborative perception by using Kalman filter\cite{kalman}, \textit{Late collaboration+Kalman Filter}. Note that SyncNet can also work as a plugin latency compensation module for other intermediate collaborative perception methods, such as V2VNet~\cite{v2vnet}. SyncNet is equivalent to feature-attention symbiotic estimation (FASE) + time modulation(TM). Corresponding to FASE with the dual-branch structure, a simplified variation is Vanilla Estimation(VE), which adopts a single-branch LSTM to estimate collaborative features only. In ablation study, we will compare the performances of \textit{DiscoNet}, \textit{DiscoNet+FASE}, \textit{DiscoNet+VE} and \textit{DiscoNet+SyncNet}.

\textbf{Training strategy.} We use a curriculum learning~\cite{curriculum} strategy in the training stage. Curriculum learning starts with easy samples and then gradually increases the difficulty. To handle the flexible latency time, we train the model under various latency settings. However, the training loss sharply increasing with the latency time causes an unstable and vulnerable training process. To tackle this issue, we employ the curriculum learning technology and gradually increase the latency time by 1 every 10 epochs until 10. Afterward, we randomly sample the latency time with an exponential distribution averaging 5 to further upgrade the model to accommodate the flexible communication latency.

\subsection{Quantitative evaluation}

Fig.~\ref{fig:comparison} compares the detection performances among our latency-aware collaborative perception system, no collaboration, DiscoNet without latency compensation, and late collaboration with a Kalman filter as a function of latency time. We see that: i) \textit{DiscoNet} is vulnerable to latency, whose performance is even lower than \textit{No Collaboration} at high latency; ii) our \textit{DiscoNet+SyncNet} is robust to latency and outperforms \textit{No Collaboration} even in a terrible communication condition with a communication latency of up to 10 frames; iii) our \textit{DiscoNet+SyncNet} consistently outperforms \textit{DiscoNet} under varying communication latency, and improves performance in AP@0.5/0.7 by up to 15.6\%/12.6\%.

Fig.~\ref{transformer} shows the performances of other frameworks, including V2VNet and a transformer-based fusion module, with and without SyncNet. The transformer-based fusion module deploys a multi-head attention architecture~\cite{transformer} to fuse the collaborative features at each spatial position.  The SyncNet module improves the performance up to 11.8\%/8.7\% in AP@0.5, respectively. It shows that various collaborative perception models are vulnerable to latency and the proposed compensation module consistently and significantly benefits those frameworks.

\subsection{Ablation study}
We first study the effect of historical frames $k$ in Fig.~\ref{k234}. We see that, $k=3$ significantly outperforms $k=2$, but $k=4$ only brings marginal benefit. The default choice in the paper is $k=3$, achieving a reasonable balance between computation efficiency and performance. 
We further validate the effectiveness of the two major components of our proposed latency compensation module (SyncNet): FASE and TM. Vanilla Estimation(VE) adopts a single-branch structure only estimating collaborative features. Fig.~\ref{tm_com} compares \textit{DiscoNet}, \textit{DiscoNet + FASE}, \textit{DiscoNet + VE} and \textit{DiscoNet + SyncNet} as a function of latency time. We see  that: i) comparing green line with blue line, our latency-aware collaborative perception system only needs a vanilla LSTM compensation module to achieve significant performance improvement in latency scenarios. ii) comparing red line with blue line, FASE architecture can improve performance in AP@0.7 metric; iii) comparing red line with yellow line, TM can improve performance when latency is low.
Table \ref{maintable} further discusses the effectiveness of the compensation model, multi-scale convolution and time modulation module at low($\tau = 1$) and high latency($\tau = 5$). We see that: i) D surpasses A, E surpasses B, F surpasses C,  reflecting FASE is consistently effective in AP@0.7 metric; ii) C surpasses B, F surpasses E, reflecting TM is consistently effective when latency is high. 

\subsection{Qualitative evaluation}

Fig.\ref{visual_car} shows the detection results of \textit{DiscoNet} without latency, \textit{DiscoNet} with latency, \textit{DiscoNet+VE} and \textit{DiscoNet + SyncNet}. Comparing (a) with (b), we see that the correctly detected vehicles in the purple box in (a) are missed or incorrectly detected in (b) due to the latency. (c) shows that the vanilla estimation (without FASE) partially compensates latency error in the blue box but fails to achieve accurate estimation in the orange box, while our SyncNet could precisely recover the true position of both vehicles, shown in purple box of (d). Plot (d) shows that SyncNet achieves the best compensation and precisely recovers the true position of vehicles.

Fig.~\ref{weightstudy} shows the attention weight of the collaboration feature from the neighbor agent in the example shown in the first row of Fig.~\ref{visual_car}. We can see that: (b), (c) both have a similar large weights in the red box, which introduce noise into the collaboration, and (d) has a small weight like (a), here to capture the truly informative area and avoid the cascading errors caused by the inaccurate feature estimation because the attention estimation branch in SyncNet under the supervision of the ground truth of collaboration attention. These qualitative results suggests the effectiveness of SyncNet.

\section{Conclusions}

We introduce latency-aware collaborative perception and propose a novel latency compensation module, SyncNet, for time-domain synchronization, which fits in existing intermediate collaboration methods. SyncNet adopts a novel  symbiotic estimation architecture, which jointly estimates intermediate features and attention weights, as well as the time modulation, which significantly improves the overall performance at low-latency range. Comprehensive quantitative and qualitative experiments show that the proposed SyncNet can improve the perception performance in the communication latency scenario and effectively address the latency issue in the collaborative perception.

\section*{Acknowledgenments}

This research is partially supported by the National Key R\&D Program of China under Grant 2021ZD0112801, National Natural Science Foundation of China under Grant 62171276, the Science and Technology Commission of Shanghai Municipal under Grant 21511100900 and CALT Grant 2021-01.


\clearpage
%
%
\bibliographystyle{splncs04}
\bibliography{LA-Coperception}
\end{document}